\documentclass[lettersize,journal]{IEEEtran}
\usepackage{amsmath,amsfonts}
\usepackage{algorithm}
\usepackage{array}
\usepackage[caption=false,font=normalsize,labelfont=sf,textfont=sf]{subfig}
\usepackage{textcomp}
\usepackage{stfloats}
\usepackage{url}
\usepackage{verbatim}
\usepackage{graphicx}
\usepackage{cite}

\usepackage{times}
\usepackage{multirow}
\usepackage{soul}
\usepackage{tabularx}
\usepackage[utf8]{inputenc}
\usepackage{amsthm}
\usepackage{mathrsfs}
\usepackage{booktabs}
\usepackage{algpseudocode} % Add this package for algorithmicx
\usepackage[switch]{lineno}
\usepackage[table]{xcolor}

\usepackage{adjustbox}

\newtheorem{definition}{Definition}
\newtheorem{template}{Prompt}

\begin{document}

\title{Large Language Model Meets Graph Neural Network in Knowledge Distillation}

\author{
Shengxiang Hu,
Guobing Zou*,
Song Yang,
Yanglan Gan,
Bofeng Zhang*,
and Yixin Chen,~\IEEEmembership{Fellow,~IEEE,}
        % <-this % stops a space
\IEEEcompsocitemizethanks{
    \IEEEcompsocthanksitem Shengxiang Hu, Guobing Zou, and Song Yang are with the School of Computer Engineering and Science, Shanghai University, Shanghai, China. Email: \{shengxianghu, gbzou, yangsong\}@shu.edu.cn.
    \IEEEcompsocthanksitem Yanglan Gan is with the School of Computer Science and Technology, Donghua University, Shanghai 201620, China. E-mail: ylgan@dhu.edu.cn
    \IEEEcompsocthanksitem Bofeng Zhang is with the School of Computer and Information Engineering, Shanghai Polytechnic University, Shanghai, China. Email: bfzhang@sspu.edu.cn.
    \IEEEcompsocthanksitem Yixin Chen is with the Department of Computer Science and Engineering, Washington University in St. Louis, St. Louis, MO 63130 USA. E-mail: chen@cse.wustl.edu.
    \IEEEcompsocthanksitem * Corresponding authors}

}

\markboth{Journal of \LaTeX\ Class Files,~Vol.~14, No.~8, August~2021}%
{Shell \MakeLowercase{\textit{et al.}}: Large Language Model Meets Graph Neural Network in Knowledge Distillation}


\maketitle

\begin{abstract}
Recent advancements in leveraging Large Language Models (LLMs) for Text-Attributed Graphs (TAGs) learning have shown significant potential, but practical deployment is often hindered by substantial computational and storage demands. Conventional Graph Neural Networks (GNNs) are more efficient but struggle with the intricate semantics embedded in TAGs. To combine the semantic understanding of LLMs with the efficiency of GNNs, we propose a novel LLM-to-GNN knowledge distillation framework, \underline{\textbf{Lingu}}istic \underline{\textbf{G}}raph \underline{\textbf{K}}nowledge \underline{\textbf{D}}istillation (\textbf{LinguGKD}), which employs TAG-oriented instruction tuning to train pre-trained LLMs as teachers and introduces a layer-adaptive contrastive distillation strategy to align node features between teacher LLMs and student GNNs within a latent space, effectively transferring the semantic and complex relational understanding from LLMs to GNNs.
Extensive experiments across various LLM and GNN architectures on multiple datasets demonstrate that LinguGKD significantly enhances the predictive accuracy and convergence rate of GNNs without requiring additional training data or model parameters. Compared to teacher LLMs, the distilled GNNs offer superior inference speed and reduced resource requirements, making them highly practical for deployment in resource-constrained environments. Furthermore, our framework demonstrates significant potential for leveraging ongoing advancements in LLM research to continuously improve GNN performance.
\end{abstract}


\section{Introduction}
Text-Attributed Graphs (TAGs) integrate structured graph data with rich textual information, providing a comprehensive representation of complex systems and encapsulating extensive knowledge. These graphs are extensively utilized across diverse domains \cite{li2022fairlp,yang2024efficient}. 
Recently, the revolutionary impact of Large Language Models (LLMs), such as ChatGPT \cite{ouyang2022training} and Llama \cite{touvron2023llama}, on natural language processing brings new opportunities and challenges to the application and research of TAGs.

While LLMs exhibit remarkable reasoning and problem-solving capabilities for complex tasks, they are not always satisfactory in every context. Studies \cite{pan2024unifying,li2024enhanced} have shown that integrating knowledge graphs can enhance the reasoning and handling capabilities of LLMs. 
Furthermore, traditional Graph Neural Networks (GNNs) \cite{velivckovic2017graph,wu2019simplifying,chen2020simple} excel at interpreting graph structures but struggle with semantic processing \cite{9415142}, especially as the complexity and volume of associated textual data increase. LLMs, with their exceptional contextual and relational understanding, offer a novel perspective in evaluating TAGs by effectively capturing the semantic nuances embedded in textual data. Integrating LLMs with GNNs addresses the semantic gap, leveraging the structural strengths of GNNs and the semantic prowess of LLMs.
These scenarios highlight the demand for and challenges in researching LLM-based graph learning approaches, which can not only amplify semantic interpretation in TAGs but also enhance the overall performance of graph-based tasks, underscoring the necessity and significance of this research direction.

Recent advancements in LLM-based graph learning focus on two main approaches: LLM as Enhancer (LaE) and LLM as Predictor (LaP) \cite{li2023survey}. The LaE approaches \cite{he2023harnessing,chen2024exploring,wei2023llmrec} enhance node embeddings in GNNs by utilizing the semantic processing capabilities of LLMs, addressing traditional GNN limitations in extracting semantic features from TAGs. In contrast, the LaP approaches \cite{wang2023can,fatemi2023talk,ye2023natural} employ LLMs directly for prediction tasks within graph-related contexts, either by adapting Transformer-based models to incorporate graph structures or by encoding graph data in natural language for inference, thus significantly improving both semantic processing and structural understanding in graph-related tasks. 
These works demonstrate the potential of LLM as a foundational model for graph learning. However, the practical application of LLMs in graph learning faces significant challenges, particularly due to their large parameter sizes, often exceeding billions\footnote{\url{https://huggingface.co/spaces/HuggingFaceH4/open_llm_leaderboard}}. This leads to high computational and storage demands, making cost-effective and widespread deployment challenging. Additionally, LLMs' extended latency during inference presents practical limitations in operational settings.

We consider whether it is possible to leverage the powerful semantic and entity relationship understanding capabilities of graph-oriented LLMs while reducing their computational and storage resource consumption, making them suitable for large-scale deployment in resource-constrained production environments. Addressing these challenges, knowledge distillation (KD) \cite{chen2022structure2,samy2023graph2feat,joshi2021gc} from LLMs to GNNs emerges as a promising strategy, enabling the transfer of LLM insights to more compact GNNs. This approach not only aims to reduce the model size and computational demands of LLMs but also leverages GNNs' strengths in processing structured graph data. By distilling the semantic and structural understanding of LLMs into lightweight GNNs, it is expected to optimize graph reasoning tasks, facilitating more effective and adaptable real-world applications. However, LLMs and GNNs are designed for different types of data with significantly different model architectures, posing substantial challenges for knowledge transfer from LLMs to GNNs. This challenge remains largely unexplored in current research. Thus, developing effective methods for knowledge distillation between these models is crucial to unlocking their combined potential.

Under the above motivations, we propose a novel LLM-to-GNN knowledge distillation framework, \underline{\textbf{Lingu}}istic \underline{\textbf{G}}raph \underline{\textbf{K}}nowledge \underline{\textbf{D}}istillation (\textbf{LinguGKD}). To the best of our knowledge, this is the first framework that directly distills knowledge from teacher LLMs to student GNNs.
Given the very early stage of graph-oriented LLM research and the lack of off-the-shelf LLMs designed specifically for graph tasks, inspired by \cite{ye2023natural}, we begin by instruction tuning a pre-trained LLM (PLM) with carefully designed, tailored graph instruction prompts. This process equips the PLM with the capability to understand and process graph structures, thus obtaining an effective teacher LLM, named \textbf{LinguGraph LLM}. Then we introduce a layer-adaptive contrastive distillation strategy, complemented by a feature alignment mechanism to synchronize the feature spaces of the LLM and GNN, ensuring that the hierarchical node features learned by the teacher LLM are effectively aligned with those extracted by the student GNN. By doing so, it can propagate the teacher LLM’s deep semantic knowledge and intricate understanding of graph structures to the student GNN, leading to better TAG understanding capabilities.

Our extensive experimental evaluations, focused on node classification tasks, span various LLM and GNN models as well as multiple benchmark datasets, demonstrating the efficacy of the proposed LinguGKD framework in distilling graph knowledge from teacher LLMs to student GNNs and verifying its strong generality, suitable for different architectures of LLM and GNN. Specifically, the distilled GNNs exhibit a significant reduction in model complexity, making them more suitable for real-world applications with a much lower parameter count compared to LLMs, and a notable increase in inference speed. From the perspective of effectiveness, distilled GNNs not only achieve higher accuracy and faster convergence rates than the vanilla versions but also outperform those with advanced designs in certain scenarios, all without the need for additional training data or architectural changes. These results validate the effectiveness of our knowledge distillation framework and underscore its potential to enhance the practicality of LLMs in graph data processing, achieving an optimal balance between performance and efficiency.

The main contributions of this paper are summarized as follows:
\begin{itemize}
\item We conceptualize the novel research problem of knowledge distillation from LLMs to GNNs and propose an innovative graph knowledge distillation framework termed \textbf{LinguGKD}, which leverages the comprehensive semantic insights of graph-oriented teacher LLMs to enrich student GNNs' structural learning capabilities while maintaining their high efficiency.

\item Within the LinguGKD framework, we develop a unique layer-adaptive contrastive distillation strategy, which ensures effective synchronization of hierarchical node features between the teacher LLM and the student GNN, guaranteeing the transfer of deep semantic knowledge and complex graph structural understanding.

\item Extensive experimental evaluations across diverse LLM and GNN models as well as multiple benchmark datasets demonstrate that LinguGKD significantly enhances the classification accuracy of student GNNs while maintaining the lightweight nature. The distilled GNNs strike an optimal balance between performance in downstream tasks and efficiency in terms of time and space, making them practical for deployment on user devices in real-world scenarios.
\end{itemize}

The remainder of this paper is organized as follows: Section \ref{sec:problem_formulation} provides the foundations and formulates the research problem; Section \ref{sec:approach} delves into the details of the LinguGKD framework; Section \ref{sec:experiments} presents extensive experimental results and analyses; Section \ref{sec:related_work} reviews relevant literature; and Section \ref{sec:conclusion} concludes the paper.

\section{Preliminaries}
\label{sec:problem_formulation}
This section formalizes the key concepts central to our study.

\begin{figure*}[t]
  \centering 
  \includegraphics[width=0.99\textwidth]{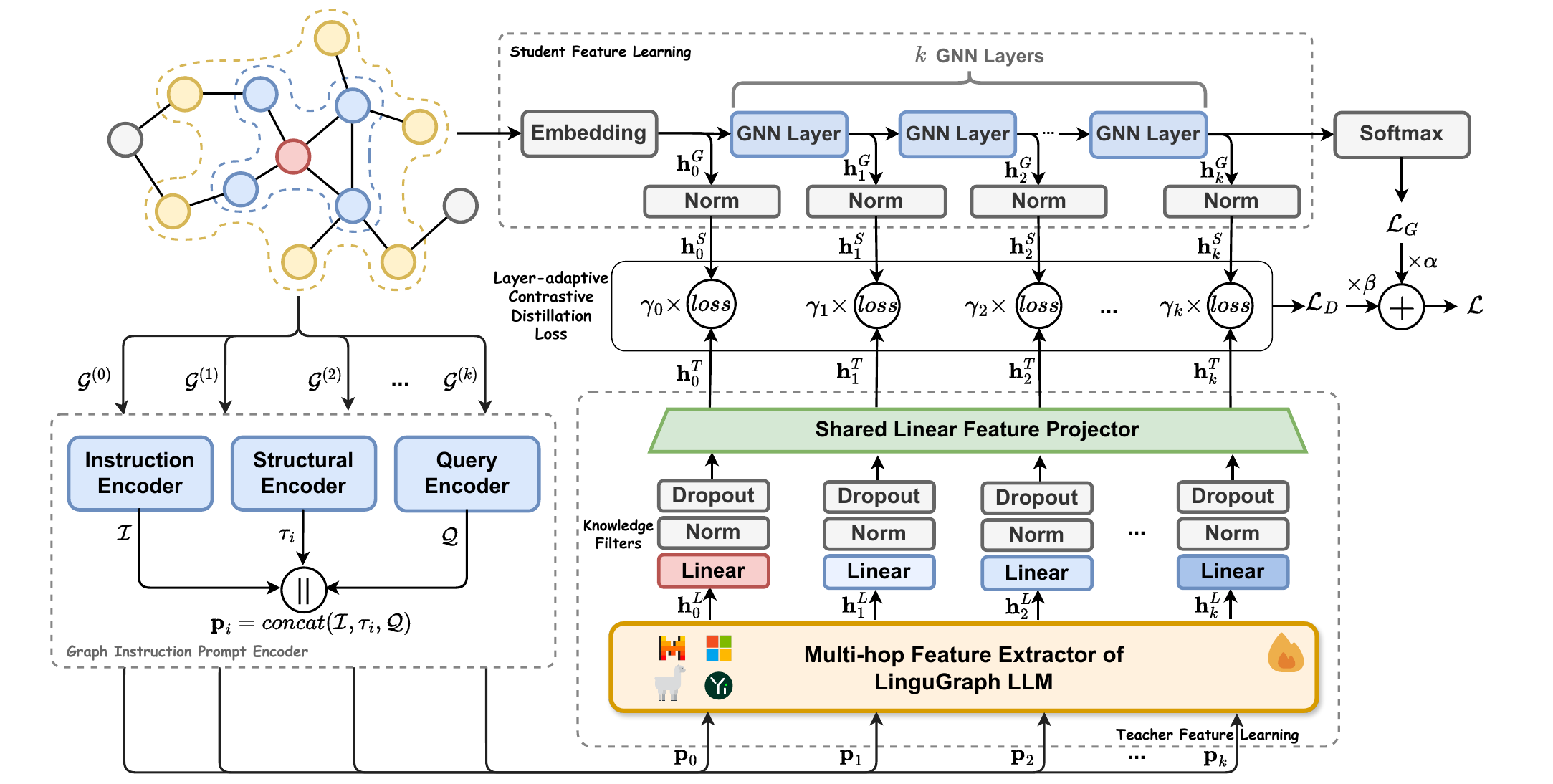}
  \caption{The proposed LinguGKD framework for TAG-oriented LLM-to-GNN knowledge distillation includes three main components: (1) Teacher feature learning using a graph-instruction-tuned LinguGraph LLM, (2) Student feature learning using a GNN, and (3) Layer-adaptive contrastive distillation to align the feature representations of both models. The lower left section depicts the process of describing the structure of a TAG with tailored graph instruction prompts for teacher feature learning.}
  \label{fig:framework}
\end{figure*}

\begin{definition}[Text-Attributed Graph]
A Text-Attributed Graph (TAG) is a graph where each node is associated with textual data. Formally, a TAG is denoted as $\mathcal{G} = (\mathcal{V},\mathcal{E}, \mathcal{X})$, where $\mathcal{V}=\{v_i\}_{i=1}^n$ represents the set of nodes, with $n$ being the total number of nodes, $\mathcal{E}$ is the set of edges with $e_{ij} \in \mathcal{E}$ indicating an edge between nodes $v_i$ and $v_j$, and $\mathcal{X}=\{x_i\}_{i=1}^n$ denotes the node attributes, where $x_i$ represents the textual attribute of node $v_i$.
\end{definition}

Given a TAG $\mathcal{G}$, GNNs are essential for interpreting the graph's topological dependencies:
\begin{definition}[Graph Neural Network]
Graph Neural Networks (GNNs) are specialized for handling graph-structured data, primarily through a $k$-layer message-passing mechanism \cite{kipf2016semi}, enabling the capture and analysis of $k$-hop node relationships and graph dynamics. This process is defined as:
\begin{equation}
\textbf{h}_v^{(k)} = f( \textbf{h}_v^{(k-1)}, \bigoplus_{u \in \mathcal{N}(v)} g(\textbf{h}_u^{(k-1)}, \textbf{h}_v^{(k-1)}))
\end{equation}
where $\textbf{h}_v^{(k)}$ is the feature representation of node $v$ at the $k$-th layer, $\mathcal{N}(v)$ includes $v$'s neighboring nodes, and functions $g(\cdot)$ and $f(\cdot)$ are trainable, responsible for aggregating neighbor features and updating node features, respectively. The operator $\bigoplus$ denotes an aggregation function, such as summation \cite{zhang2020cfgnn} or averaging \cite{kipf2016semi}.
\end{definition}

While GNNs excel at processing graph structures, they have limitations in understanding individual node semantics. To leverage the capability of LLMs in semantic feature learning, an LLM-to-GNN Graph Knowledge Distillation framework can be designed to distill knowledge from a teacher LLM to a student GNN, enhancing the GNN's capability in interpreting node semantics and complex graph topology. This can be defined as follows:

\begin{definition}[LLM-to-GNN Graph Knowledge Distillation]
An LLM-to-GNN Graph Knowledge Distillation (GKD) framework can be formulated as a quintuple: $\langle\mathcal{M}_T, \mathcal{M}_S, \mathcal{S}, \mathcal{F}, \mathcal{A}\rangle$, where the teacher model $\mathcal{M}_T$ is a Transformer \cite{vaswani2017attention}-based LLM, fine-tuned on TAG $\mathcal{G}$ for generative graph inference tasks; $\mathcal{M}_S$ denotes the student GNN specified for discriminative tasks; $\mathcal{F}$ represents the knowledge to be transferred from $\mathcal{M}_T$ to $\mathcal{M}_S$ via distillation scheme $\mathcal{S}$, learned by extraction algorithm $\mathcal{A}$. The distillation process can be formulated as follows:
\begin{gather}
    \mathcal{F}=\{\mathcal{F}_T, \mathcal{F}_S\} = \{\mathcal{A}_T(\mathcal{M}_T,\mathcal{G}), \mathcal{A}_S(\mathcal{M}_S,\mathcal{G})\}\\
    \mathcal{L}_{D} = \text{loss}(\mathcal{F}_S, \mathcal{F}_T) \\
    \mathcal{M}_S^{\text{new}} = \mathcal{S}(\mathcal{M}_S, \mathcal{L}_{D})
\end{gather}
where $\mathcal{A}_T$ and $\mathcal{A}_S$ are the knowledge extraction algorithms of $\mathcal{M}_T$ and $\mathcal{M}_S$, respectively; $\text{loss}(\cdot)$ denotes the divergence function (e.g., Kullback-Leibler divergence), and $\mathcal{M}_S^{\text{new}}$ is the distilled student model that we ultimately require.
\end{definition}

\section{Approach}
\label{sec:approach}
Figure \ref{fig:framework} illustrates the LinguGKD framework for TAG-oriented graph knowledge distillation, highlighting three key components: teacher feature learning by the LLM, student feature learning by the GNN, and layer-adaptive contrastive distillation loss between the two feature sets. Before delving into these crucial modules, we elaborate on how to fine-tune a TAG-oriented Pre-trained Language Model (PLM) to understand graph structure and node semantics through our tailored instruction prompts.

\subsection{TAG Instruction Tuning of Pre-trained LLM}
Given a TAG $\mathcal{G}=(\mathcal{V}, \mathcal{E}, \mathcal{X})$, a center node $v_i$ and a specific neighbor hop $k$, we can obtain a collection of neighbor subgraphs from structural-free up to the $k$-th hop: $\mathcal{G}_i^k=\{\mathcal{G}_i^{(l)}\}_{l=0}^k$, where $\mathcal{G}_i^{(k)}=(v_i, \mathcal{N}_i^{(k)}, \mathcal{E}_{\mathcal{N}_i^{(k)}}, \mathcal{X}_{\mathcal{N}_i^{(k)}})$ denotes the $k$-th hop subgraph, in which $\mathcal{N}_i^{(k)}$ is the set of $k$-hop neighbors of $v_i$, and $\mathcal{E}_{\mathcal{N}_i^{(k)}}$, $\mathcal{X}_{\mathcal{N}_i^{(k)}}$ are the corresponding edges and textual node attributes within the subgraph, respectively.

To enable the PLM to accurately understand graph structure and node semantics, it is essential to craft comprehensive instruction prompts for further instruction tuning. For a given $\mathcal{G}_i^{(k)}$, we define a specific instruction prompt $\textbf{p}_k$, which consists of three components: task-specific instructions $\mathcal{I}$ that delineate the expected model action, the structural prompt $\tau_k$ that is the natural language description of the subgraph, and a task-relevant query $\mathcal{Q}$, typically presented as a detailed question. The construction of $\textbf{p}_k$ is a concatenation of these elements, as shown in the lower left part of Figure \ref{fig:framework}:
\begin{equation}
\textbf{p}_k = \text{concat}(\mathcal{I}, \tau_k, \mathcal{Q})
\end{equation}

Here, we employ node classification as a side task to enable the LLM to comprehend graphs. In alignment with OpenAI's Prompt Engineering principles \cite{OpenAIPromptEngineering}, we carefully design $\mathcal{I}$ and $\mathcal{Q}$ as follows:

\begin{template}[TAG Node Classification Instruction]
Implement a node classification system for \textbf{\{\{type of graph\}\}}, representing nodes as tuples (node\_\{id\}, \{degree\}, \{attribute\}). Classify nodes into [\textbf{\{\{list of classification categories\}\}}] based on attributes and link relations. \textbf{\{\{classification criteria for a specific graph\}\}}.
\end{template}

\begin{template}[TAG Node Classification Query]
Which category should (node\_\textbf{\{\{id\}\}}, \textbf{\{\{node degree\}\}}, \textbf{\{\{node attributes\}\}}) be classified as?
\end{template}

In these templates, each node is represented by a tuple that encapsulates the node's id, degree, and the corresponding textual attributes. The placeholders within curly braces \{\{\}\} are filled based on the specific graph data. More precisely, the term \textbf{\textit{\{\{type of graph\}\}}} specifies the domain this graph belongs to, while \textbf{\textit{\{\{list of classification categories\}\}}} enumerates the potential categories for node classification. Furthermore, \textit{\textbf{\{\{classification criteria for a specific graph\}\}}} denotes the optional prior knowledge that assists the LLM in the precise generation of labels for the center node.

In crafting the structural prompt, we prioritize key elements, such as node textual attributes, node degrees, and multi-order neighbor interactions, aligned with conventional multi-layer message-passing GNNs' principles. Referring to the strategies from \cite{ye2023natural}, we develop a linguistic structural encoder $f_e$ that transforms $\mathcal{G}_{i}^{(k)}$ into a detailed natural language description, represented as $\tau_k$:
\begin{equation}
\tau_k = f_e(\mathcal{G}_{i}^{(k)})
\end{equation}

The prompt template of $f_e$ is meticulously designed as follows:
\begin{template}[TAG Structural Prompt]
\label{temp:graph}
(node\_\textbf{\{\{id\}\}}, \textbf{\{\{node degree\}\}}, \textbf{\{\{node attributes\}\}}) is connected within \textbf{\{\{k\}\}} hops to \textbf{\{\{k-th hop neighbors\}\}} through paths that may involve \textbf{\{\{intermediate paths\}\}}.
\end{template}

In this context, \textbf{\textit{\{\{k-th hop neighbors\}\}}} refers to the ensemble of neighbors reachable at the $k$-th hop in $\mathcal{N}_i^{(k)}$. Each neighbor in this list is characterized by a tuple that encapsulates its id, degree, and attributes, mirroring the representation of the central node. Furthermore, \textbf{\textit{\{\{intermediate paths\}\}}} represents the sequences of paths connecting the central node $v_i$ to its $k$-th hop neighbors, encompassing the edges defined within $\mathcal{E}_{\mathcal{N}_i^{(k)}}$. This template meticulously outlines the connectivity and relational dynamics within the subgraph based on node degrees, attributes, and the specified hop distances.

Iterating the structural encoding process, we can methodically generate a structural prompt for every subgraph within $\mathcal{G}_i^k$, leading to the assembly of the $k$-hop structural prompt set $\mathcal{T}=\{f_e(\mathcal{G}_i^{(l)})\}_{l=0}^k$ for the central node $v_i$. Notably, $l=0$ focuses exclusively on $v_i$'s textual attributes. This iterative generation of structural prompts systematically captures the nuanced relational dynamics at varying degrees of connectivity, from the immediate vicinity ($l=0$) extending to the broader $k$-hop network. By concatenating instruction $\mathcal{I}$ and query $\mathcal{Q}$ with various structural prompts, we can obtain a set of graph instruction prompts $\mathcal{P}$:
\begin{equation}
\mathcal{P}=\{\text{concat}(\mathcal{I}, \tau_l, \mathcal{Q})\}, \quad \forall \tau_l \in \mathcal{T}
\end{equation}

Then we adopt instruction tuning \cite{zhang2023instruction, si2023instruction, honovich2022unnatural} to specifically tailor PLMs for generative node classification tasks. Specifically, each prompt $\textbf{p}_l \in \mathcal{P}$ is utilized to directly fine-tune the PLM to generate the semantic category $\textbf{y}_i$ for the corresponding center node, without any modification of the pre-trained tokenizer and vocabulary. We utilize negative log-likelihood loss as our objective function:
\begin{equation}
\begin{aligned}
\mathcal{L}_{T}(\mathcal{P}) = -\sum_{\textbf{p}_l \in \mathcal{P}} \sum_{j=1}^{|\textbf{y}|} \log p(\hat{y}_j|\textbf{p}_l, \hat{y}_{<j})
\end{aligned}
\end{equation}
where $\hat{y}_j$ refers to the $j$-th token that the LLM generated for the node label. Employing this instruction tuning methodology, we develop a highly proficient LLM adept in TAG understanding. This fine-tuned LLM, which we call LinguGraph LLM, then acts as the teacher model $\mathcal{M}_T$ in the subsequent graph knowledge distillation process.

\subsection{Knowledge Distillation from LinguGraph LLM to GNN}
In this section, we detail the key components of the proposed LinguGKD framework, addressing the challenge of transferring insights from complex teacher LLMs to simpler student GNNs. This process emphasizes aligning the node latent features extracted by both the LLM and the GNN within a unified latent space. The following subsections elaborate on three pivotal phases of LinguGKD's knowledge distillation process: extracting semantically-enhanced node features via LLMs, leveraging GNNs for structural node feature extraction, and implementing layer-adaptive alignment of semantic and structural features to distill knowledge from the teacher LLM to the student GNN.

\subsubsection{\textbf{Teacher Feature Learning via LinguGraph LLM}}
We begin by leveraging the fine-tuned LinguGraph LLM $\mathcal{M}_T$ to extract semantically-enriched node features, encapsulating textual attributes and multi-order neighbor information, as depicted in the Teacher Feature Learning module in Figure \ref{fig:framework}.

Building upon insights from \cite{bge_embedding}, we observe that tailored instructions significantly enhance the LLM's proficiency in generating semantic features. Consequently, we leverage the entire instruction prompt set $\mathcal{P}$ for the extraction of node semantic features, rather than limiting it to the structural prompt set $\mathcal{T}$. Specifically, given that LLMs primarily use the transformer architecture \cite{vaswani2017attention}, which includes an embedding layer for token embedding, an $n$-layer transformer with multi-head self-attention for deriving word-level nonlinear interrelations, and an output layer for specific generative tasks, we extract node semantic features as follows.

For an instruction prompt $\textbf{p}_l \in \mathcal{P}$, we process the sequence of tokens through the embedding layer:
\begin{equation}
E^L = Embedding^L(\{\rho_i\}_{i=1}^{|\textbf{p}_l|}; W_{\text{emb}}^L)
\end{equation}
where $Embedding^L(\cdot)$ is the embedding layer of the LLM $\mathcal{M}_T$, and $W_{\text{emb}}^L$ represents its parameters. $E^L$ is the embedded sequence, each row of which represents a token.

Next, the embedded sequence is fed into the transformer layers:
\begin{equation}
H = Transformer(E^L; W_{\text{tr}})
\end{equation}
where $Transformer(\cdot)$ denotes the transformer module of LLM $\mathcal{M}_T$, and $W_{\text{tr}}$ represents its parameters. The transformer layers, using multi-head self-attention, compute the contextual relationships between tokens, resulting in a contextualized feature matrix $H$.

Finally, we consider the feature of the last token $\rho_{|\textbf{p}_l|}$ from the final transformer layer as the $l$-th order node latent feature:
\begin{equation}
\textbf{h}_{l}^L = H_{|\textbf{p}_l|,:}
\end{equation}

Here, $\textbf{h}_{l}^L \in \mathbb{R}^{d_L}$ encapsulates the aggregated contextual information of the entire instruction prompt, integrating neighbor details and extensive node attribute data, where $d_L$ denotes the dimension of the node latent features extracted by the teacher LLM. Through this process, we obtain a rich semantic representation for each node, leveraging the comprehensive understanding capabilities of the LLM.

For distilling hierarchical knowledge from the teacher LLM, we then process each $l$-th order feature, $\textbf{h}_l^L$, through a specialized hop-specific neural knowledge filter, $\mathcal{M}_{f}^l$, which is proficient in filtering pertinent layer knowledge:
\begin{equation}
\mathcal{M}_{f}^l(\textbf{h}_l^L) = \sigma(W_l \textbf{h}_l^L + b_l), \quad 0 \le l \le k
\end{equation}
\begin{equation}
\hat{\textbf{h}}_l^L = LayerNorm(\mathcal{M}_{f}^l(\textbf{h}_l^L))
\end{equation}
where $W_l, b_l$ are the trainable parameters for the filter $\mathcal{M}_{f}^l$, and $\sigma$ denotes the non-linear activation function.

Subsequently, to align the node features from different hops into the same lower-dimensional distillation vector space without altering their distribution, a cross-hop shared linear feature projector, $\mathcal{M}_{p}$, is designed to restructure these features:
\begin{equation}
\textbf{h}_l^T = \mathcal{M}_{p}(\hat{\textbf{h}}_l^L; W_{p}, b_{p}) = W_{p} \hat{\textbf{h}}_l^L + b_{p}
\end{equation}
where $\textbf{h}_l^T \in \mathbb{R}^{d_k}$ symbolizes the adapted $l$-order teacher knowledge, poised for subsequent distillation steps; $W_p, b_p$ are the trainable parameters for $\mathcal{M}_p$.

Applying this process across all subgraph orders up to the $k$-th, we obtain a set of hierarchical teacher node features $\mathcal{F}_T = \{\textbf{h}_l^T\}_{l=0}^k$, which is then utilized in the subsequent graph knowledge distillation, serving as the specific knowledge to be distilled from the teacher LLM to the student GNN.

\subsubsection{\textbf{Student Feature Learning via GNN}}
We then leverage the student GNN $\mathcal{M}_S$ to extract multi-hop node features, as shown in the Student Feature Learning module in Figure \ref{fig:framework}. The chosen student model can be any off-the-shelf GNN. Although various GNN models have different architectures \cite{kipf2016semi,velivckovic2017graph,hamilton2017inductive,xu2018powerful}, they all interpret graph structures through a message-passing mechanism that enables central nodes to assimilate features from $k$-hop neighbors, capturing local graph structure nuances. The core principle of message-passing remains consistent: a structured process involving message construction, aggregation, and node feature updating. These stages allow GNNs to effectively represent intricate graph structures.

For a given $k$-hop neighbor subgraph $\mathcal{G}_i^{(k)}$ of a central node $v_i$, the $k$-order message aggregation process in GNN unfolds as follows:
\begin{gather}
\textbf{h}_j^{(0)} = Embedding^G(x_j;W_{\text{emb}}^G), \quad \forall v_j \in v_i \cup \mathcal{N}_i^{(k)} \\
\textbf{m}_{i\leftarrow j}^{(l)} = \mathcal{M}^{(l)}_{msg}(\textbf{h}_i^{(l-1)}, \textbf{h}_j^{(l-1)}, e_{ij}; W_{msg}^{(l)}), \quad 0<l\le k \\
\textbf{h}_l^G = \mathcal{M}_{update}^{(l)}(\textbf{h}_i^{(l-1)}, \bigoplus_{v_j\in \mathcal{N}_{i}}\textbf{m}_{i\leftarrow j}^{(l)} ; W_{update}^{(l)})
\end{gather}
where $x_j \in \mathcal{X}_{\mathcal{N}_i^{(k)}}$ represents the attribute of node $v_j$, $Embedding^G(\cdot)$ is the text embedding model of $\mathcal{M}_S$, such as bag-of-words or TF-IDF, converting node textual attributes into a low-dimensional vector space, establishing the initial node feature $\textbf{h}_j^{(0)}$. The functions $\mathcal{M}^{(l)}_{msg}(\cdot)$ and $\mathcal{M}^{(l)}_{update}(\cdot)$ perform message construction and node feature updates, respectively, during $l$-order message passing. $\bigoplus$ denotes a differentiable, permutation-invariant function (e.g., sum, mean, or max) to perform message aggregation. The parameters $W_{msg}^{(l)}$ and $W_{update}^{(l)}$ are the associated trainable weights. The output of the $l$-th message passing layer $\textbf{h}_l^{G} \in \mathbb{R}^{d_G}$ represents the $l$-th order feature of the central node $v_i$, reflecting its $l$-order neighbor structure and attributes.

Subsequently, we synchronize the node features into the unified distillation vector space via a normalization layer:
\begin{equation}
\textbf{h}_l^S = Norm(\textbf{h}_{l}^G), \quad 0 \le l \le k 
\end{equation}
where $\textbf{h}_l^S \in \mathbb{R}^{d_k}$ denotes the student knowledge, and $Norm(\cdot)$ is a normalization function, typically batch or layer normalization. Applying this process across all subgraph orders up to the $k$-th, we obtain a set of hierarchical student node features $\mathcal{F}_S = \{\textbf{h}_l^S\}_{l=0}^k$. 

In the next phase, we focus on aligning the student GNN's learned node features $\mathcal{F}_S$ with those from the teacher LLM, $\mathcal{F}_T$. This alignment ensures that the LLM's rich semantic knowledge is effectively integrated into the GNN, enhancing its capability to process graph-structured data.

\subsubsection{\textbf{Layer-Adaptive Contrastive Distillation}}
\label{subsec:distill}
In graph inference, the relevance of node features varies significantly with the order of their neighbors: structural-free features highlight a node's core attributes, lower-order features elucidate direct interactions, and higher-order features provide insight into distant connections, which are essential for understanding a node's role within the graph. In different downstream tasks, the importance of features encoding information from various neighbor orders can differ significantly. For instance, tasks such as community detection may rely more on higher-order features \cite{huang2019higher}, while node classification might benefit more from lower-order features \cite{kipf2016semi}.

Given these variations, a one-size-fits-all distillation approach may fail to capture the nuanced importance of features at different hops. Therefore, to fully leverage the teacher LLM's comprehensive understanding of different neighbor orders, we propose a Layer-Adaptive Contrastive Distillation mechanism within the LinguGKD framework. This approach tailors the distillation process to the importance of node features at each hop, ensuring the student GNN effectively captures the teacher LLM's nuanced knowledge of both local and global graph structures. This layer-adaptive strategy is essential for optimizing the student GNN's performance across diverse tasks by facilitating precise and task-relevant knowledge transfer.

To achieve effective knowledge transfer, it is crucial to align the feature spaces of the teacher LLM and the student GNN. Contrastive learning with infoNCE loss \cite{oord2018representation} is particularly well-suited for this task because it encourages the alignment of similar (positive) feature pairs while ensuring that dissimilar (negative) pairs are distinguishable. By leveraging contrastive learning, we can effectively measure and minimize the divergence between the layer-wise features extracted by both models, thereby ensuring that the student GNN can accurately mimic the teacher LLM's deep semantic understanding. 

To design an effective contrastive distillation loss, we consider the following steps: For each node $v_i$ and its corresponding $l$-th order feature, the positive sample is the same node's feature as learned by both the teacher LLM and the student GNN. This ensures that the features of the same node are aligned across different models. 
The negative samples are features of nodes from different categories compared to the center node $v_i$. This selection helps in maintaining a clear distinction between different classes, thereby improving the classification accuracy of the student GNN. 
Formally, given a node $v_i$, its $l$-th order feature from the teacher LLM, $\textbf{h}_l^T$, and from the student GNN, $\textbf{h}_l^S$, the positive pair is defined as:
\begin{gather}
    (\textbf{h}_l^S, \textbf{h}_l^T)
\end{gather}

For each positive pair, we sample $N$ negative pairs $\{(\textbf{h}_l^S,\textbf{h}_l^{T*})\}$, where $\textbf{h}_l^{T*}$ is the $l$-th order feature of a node from a different category.
The infoNCE loss for each $l$-th order feature, controlled by a temperature parameter $t$, is expressed as follows:
\begin{gather}
    \mathcal{L}_{D}^l = -\mathbb{E} \left[ \log \frac{\exp(\text{sim}(\textbf{h}_l^S, \textbf{h}_l^T)/t)}{\sum_{m=1}^{N}\exp(\text{sim}(\textbf{h}_l^S, \textbf{h}_{l,(m)}^{T*})/t)} \right]
\end{gather}
where $sim(\cdot,\cdot)$ denotes a similarity function, such as cosine similarity. The temperature parameter $t$ controls the smoothness of the probability distribution, ensuring that the model focuses on hard negative samples that are more challenging to distinguish from the positives.

Recognizing the unique importance of different-order neighbor structures in various downstream tasks, we then introduce a trainable distillation factor $\gamma_l$ for each layer's distillation loss, which allows the model to adaptively focus more on the layers that are more critical for the specific task at hand. 
The overall layer-adaptive contrastive distillation loss is then computed as the weighted sum of the layer-specific contrastive losses:
\begin{equation}
    \mathcal{L}_{\text{D}} = \sum_{l=0}^k \gamma_l \mathcal{L}_{\text{D}}^l
\end{equation}

Here, each order's distillation factor $\gamma_l$ ensures a balanced knowledge distillation, enabling the effective transfer of complex semantic and structural insights from the LLM to the GNN.

\subsection{Model Training}
In training the student GNN, acknowledging the different inference frameworks of the teacher and student models, it is essential to not only distill knowledge from the teacher LLM but also to train a task-specific prediction layer for the GNN. Therefore, we approach the student GNN's training as a multi-task joint optimization challenge. For instance, in the node classification scenario, we use a fully connected layer as the classifier:
\begin{equation}
    \hat{y} = softmax(W_{\text{G}} \textbf{h}_k^S + b_{\text{G}})
\end{equation}
where $\textbf{h}_k^S$ denotes the output of the $k$-th layer of the GNN, $\hat{y}$ represents the predicted node label, and $W_{\text{G}}$ and $b_{\text{G}}$ are the weights and biases of the fully connected layer.

Subsequently, the node classification loss function, formulated as cross-entropy, is computed as:
\begin{equation}
    \mathcal{L}_{\text{G}} = -\sum_{i=1}^{|\mathcal{D}_{\text{tr}}|} y_i \log(\hat{y}_i)
\end{equation}
where $y_i$ denotes the actual node label, and $\hat{y}_i$ is the predicted probability for each category. $\mathcal{D}_{\text{tr}}$ is the training set.

The overall training objective integrates the KD loss $\mathcal{L}_{\text{D}}$ with the classification loss $\mathcal{L}_{\text{G}}$, obtaining a joint loss function:
\begin{equation}
    \mathcal{L} = \alpha \mathcal{L}_{\text{G}} + \beta \mathcal{L}_{\text{D}}
\end{equation}
where $\alpha$ and $\beta$ are tunable factors for adaptively balancing the influence of knowledge distillation loss and downstream task loss on the training process.

Finally, the student GNN undergoes end-to-end training using a mini-batch AdamW optimizer, which is dedicated to optimizing model parameters efficiently for robust performance.

\section{Experiments}
\label{sec:experiments}

\begin{table}[t]
\centering
\caption{Dataset Statistics}
\label{tab:dataset}
\begin{tabularx}{0.47\textwidth}{lXXX}
\toprule
& \textbf{Cora} & \textbf{PubMed} & \textbf{Arxiv} \\
\midrule
\# Node & 2,708 & 19,717 & 169,343 \\
\# Edge & 5,429 & 44,338 & 1,166,243 \\
\# Class & 7 & 3 & 40 \\
\# Features & 1433 & 500 & 128 \\
Embedding Tech. & BoW & TF-IDF & Skip-gram \\
$|\mathcal{D}_{tr}|:|\mathcal{D}_{val}|:|\mathcal{D}_{test}|$ & 6:2:2 & 6:2:2 & 5.4:1.8:2.8 \\
\bottomrule
\end{tabularx}
\end{table}

\subsection{Datasets and Backbone Models}
\paragraph{\textbf{Datasets}}
We validated the effectiveness of our proposed LinguGKD framework through node classification experiments on three widely-adopted benchmark datasets: Cora, PubMed \cite{yang2016revisiting}, and Arxiv \cite{hu2020open}. These datasets represent academic papers as nodes and citations as edges. The node attributes consist of titles and abstracts, encapsulating the core content of each paper. For knowledge distillation and GNN training, we utilized the default node embeddings generated by various techniques (e.g., bag of words (BoW), TF-IDF) inherent to these datasets without any alterations. 

Due to the lack of initial text attributes for each node in the original datasets, we reconstructed titles, abstracts, and other text attributes for each node following the method described in \cite{he2023harnessing} for graph instruction tuning of the teacher LLM.

For dataset partitioning, we followed the split strategy adopted in \cite{ye2023natural,he2023harnessing}. Specifically, we applied a 6:2:2 split for the Cora and PubMed datasets, while for the Arxiv dataset, we used the 5.4:1.8:2.8 split as in the OGB open benchmark \cite{hu2020open}. Comprehensive dataset statistics are summarized in Table \ref{tab:dataset}. 

\paragraph{\textbf{Backbone Models}}
To verify the effectiveness and generality of our proposed LinguGKD framework, we selected multiple LLMs with different architectures as teachers and GNNs as students. Specifically, we chose Mistral-7B \cite{jiang2023mistral}, Llama2-7B \cite{touvron2023llama}, and Llama3-8B \cite{llama3modelcard} as the teacher models. For student GNNs, we selected GCN \cite{hamilton2017inductive}, GAT \cite{velivckovic2017graph}, GraphSAGE \cite{hamilton2017inductive}, and GIN \cite{xu2018powerful}, known for their effectiveness and efficiency in graph-based tasks. 

\subsection{Experimental Settings}

\begin{table}[t]
\centering
\caption{Overview of Experimental Settings}
\label{tab:exp_settings}
\begin{tabularx}{\linewidth}{Xl}
\toprule
\textbf{Parameter} & \textbf{Value} \\
\midrule
\multicolumn{2}{c}{\textbf{Instruction Tuning of PLM}} \\
Maximum neighboring subgraph order ($k$) & 3 \\
Maximum number of instruction prompts per hop ($\theta$) & 2 \\
% Total instruction prompts & 62,216 (Cora), 121,580 (PubMed), 2,000,123 (Arxiv) \\
Maximum sequence length ($s_{max}$) & 512 \\
Hidden feature dimension ($d_L$) & 4096 \\
Batch size ($mb_L$) & 2 \\
Gradient accumulation ($g_{acc}$) & 2 \\
Learning rate ($lr_L$) & $2 \times 10^{-4}$ \\
Training duration ($ep_L$) & 1 epoch \\
LoRA parameters & $lora_r$=64, $lora_{\alpha}$=32 \\
% Trainable parameters & Approx. 1,800,000 \\
% Node label generation & Greedy search, majority voting \\
\midrule
\multicolumn{2}{c}{\textbf{Graph Knowledge Distillation}} \\
% GNN architectures & Various, implemented through PyG \\
Learning rate ($lr_G$) & $1 \times 10^{-4}$ \\
Batch size ($mb_G$) & 32 \\
Training epochs ($ep_G$) & 500 \\
Feature dimension ($d_G$, $d_k$) & [64, 128, 256, 512, 1024] \\
% Instruction prompts per node & $\theta$ unique prompts per hop \\
% Feature extraction & $\theta$ $l$-hop features, average pooling \\
\bottomrule
\end{tabularx}
\end{table}

\paragraph{\textbf{Instruction Tuning of PLM}}
During the LLM's fine-tuning phase, for each center node $v_i$, we defined a maximum neighboring subgraph order $k$ of 3, constructing instruction prompts $\mathcal{P}=\{\mathcal{P}_l\}_{l=0}^{3}$ to cover structural prompts from structure-free to 3rd-hop subgraphs. We standardized the maximum number of instruction prompts per hop ($\theta$) at 2, resulting in totals of 62,216, 121,580, and 2,000,123 instruction prompts for Cora, PubMed, and Arxiv, respectively.

We utilized pre-trained models from \textit{huggingface.co}. The LLM input's maximum sequence length ($s_{max}$) was capped at 512, with a hidden feature dimension ($d_L$) of 4096. Training settings included a minimal batch size ($mb_L$) of 2, gradient accumulation ($g_{acc}$) also set at 2, and a learning rate ($lr_L$) of $2 \times 10^{-4}$, all within a concise training duration of 1 epoch ($ep_L$). We integrated LoRA and 4-bit quantization techniques, with LoRA's parameters set to $lora_r$=64, dropout at 0.1, and alpha ($lora_{\alpha}$) to 32. This adjustment resulted in a total of approximately 1,800,000 trainable parameters, including elements like \textit{q\_proj}, \textit{k\_proj}, \textit{v\_proj}, \textit{o\_proj}, and \textit{lm\_head}.

In the node label generation steps of LinguGraph LLM, each node underwent classification for every instructional prompt via a greedy search algorithm. A majority voting scheme was then applied, where the most frequently appearing classification across prompts was selected as the final prediction, ensuring a balanced and democratically derived outcome.

\paragraph{\textbf{Graph Knowledge Distillation}}
For distilling graph knowledge from LinguGraph LLM to GNNs, across all benchmark datasets, we standardized the learning rate ($lr_G$) at $1 \times 10^{-4}$, batch size ($mb_G$) at 32, and extended the training to 500 epochs ($ep_G$). The performance of our LinguGKD was assessed across different message-passing layers [0, 1, 2, 3].

For simplicity, we aligned the feature dimension of the GNN ($d_G$) with that of the distilled knowledge ($d_k$). We then explored the effects of varying the hidden feature dimensions of GNNs and distilled knowledge at [64, 128, 256, 512, 1024].

For each central node $v_i$ within a specified neighbor order $l$, our experimental setup involved generating $\theta$ unique instruction prompts. This led to the extraction of $\theta$ $l$-hop features $\{\textbf{h}_{l,(1)}^L, \dots, \textbf{h}_{l,(\theta)}^L\}$ from the LinguGraph LLM for $v_i$. During the layer-adaptive knowledge distillation phase, we utilized an average pooling operation to consolidate these features into a single representation $\textbf{h}_{l}^L$ that accurately reflects the $l$-th order neighborhood's characteristics.
The overall experimental settings are summarized in Tabel \ref{tab:exp_settings}.

\begin{table*}[t]
\centering
\caption{Node classification performance of various graph learning models across selected datasets, with the highest-performing outcomes in \textit{bold}, second-best scores highlighted in gray, and best baseline performances \underline{underlined}. The prefix $LinguGraph-$ represents teacher LLMs obtained by fine-tuning different PLMs with graph instruction prompts, while GNNs with different subscripts represent student models distilled from the corresponding teacher LLMs indicated by the subscripts. The term \textit{Avg. Dist. Gains.} refers to the average knowledge distillation gains obtained by different student GNNs.}
\begin{tabularx}{\textwidth}{lXXXXlXX}
    \toprule
\multirow{2}{*}{\textbf{Methods}} & \multicolumn{2}{c}{\textbf{Cora}} & \multicolumn{2}{c}{\textbf{PubMed}} & \multirow{2}{*}{\textbf{Methods}} & \multicolumn{2}{c}{\textbf{Arxiv}}  \\ \cmidrule(lr){2-5} \cmidrule(l){7-8} 
 & \textbf{Acc.$\uparrow$} & \textbf{F1$\uparrow$} & \textbf{Acc.$\uparrow$} & \textbf{F1$\uparrow$} &  & \textbf{Acc.$\uparrow$} & \textbf{F1$\uparrow$} \\ \midrule
GCN \cite{kipf2016semi} & 86.53$\pm$0.92 & 85.66$\pm$0.78 & 86.12$\pm$0.93 & 85.64$\pm$0.82 & GCN \cite{kipf2016semi} & 71.74$\pm$0.21 & 71.04$\pm$0.37 \\
GAT \cite{velivckovic2017graph} & 86.12$\pm$0.95 & 85.05$\pm$0.88 & 85.49$\pm$0.76 & 84.89$\pm$0.71 & GAT \cite{velivckovic2017graph}& 73.66$\pm$0.33 & 72.44$\pm$0.19 \\
GraphSAGE \cite{hamilton2017inductive} & 87.08$\pm$0.85 & 85.96$\pm$0.73 & 87.69$\pm$0.92 & 87.38$\pm$0.68 & GraphSAGE \cite{hamilton2017inductive} & 71.19$\pm$0.26 & 70.87$\pm$0.45 \\
GIN \cite{xu2018powerful} & 86.60$\pm$0.91 & 85.37$\pm$0.74 & 85.84$\pm$0.92 & 85.31$\pm$0.63 & GIN \cite{xu2018powerful} & 71.62$\pm$0.47 & 71.13$\pm$0.33 \\
SGC-v2 \cite{wu2019simplifying} & 85.48$\pm$1.48 & 85.04$\pm$0.69 & 85.36$\pm$0.52 & 84.96$\pm$0.77 & DeeperGCN \cite{li2020DeeperGCN} & 71.92$\pm$0.16 & 71.24$\pm$0.39 \\
BernNet \cite{he2021bernnet} & 88.52$\pm$0.95 & 87.96$\pm$0.85 & 88.48$\pm$0.41 & 87.52$\pm$0.79 & GTAN \cite{wu2022gtnet} & 72.97$\pm$0.17 & 71.77$\pm$0.22 \\
FAGCN \cite{bo2021beyond} & 88.85$\pm$1.36 & 87.92$\pm$0.65 & 89.98$\pm$0.54 & 88.72$\pm$0.53 & UniMP \cite{shi2020masked} & 73.11$\pm$0.20 & 72.14$\pm$0.38 \\
GCNII \cite{chen2020simple} & 88.93$\pm$1.37 & 87.58$\pm$0.71 & 89.80$\pm$0.30 & 88.96$\pm$0.62 & GCNII \cite{chen2020simple} & 72.74$\pm$0.00 & 72.22$\pm$0.44 \\
RevGAT \cite{li2021training} & 89.11$\pm$0.00 & 87.65$\pm$0.58 & 88.50$\pm$0.05 & 87.12$\pm$0.73 & RevGAT \cite{li2021training} & \underline{74.02$\pm$0.18} & \underline{73.56$\pm$0.29} \\
ACM-GCN+ \cite{luan2022Revisiting} & \underline{89.75$\pm$1.16} & \underline{88.94$\pm$0.54} & \cellcolor[gray]{0.8}\underline{90.96$\pm$0.62} & \underline{89.77$\pm$0.51} & E2EG \cite{dinh2023e2eg} & 73.62$\pm$0.14 & 72.96$\pm$0.26 \\
GraphTransformer \cite{dwivedi2020generalization} & 86.42$\pm$0.82 & 85.96$\pm$0.67 & 88.75$\pm$0.16 & 87.91$\pm$0.59 & SGFormer \cite{wu2024simplifying} & 72.63$\pm$0.13 & 71.58$\pm$0.42 \\ 
Graphormer \cite{ying2021transformers} & 80.41$\pm$0.30 & 79.98$\pm$0.56 & 88.24$\pm$1.50 & 87.52$\pm$0.71 & Graphormer \cite{ying2021transformers} & 72.81$\pm$0.23 & 72.14$\pm$0.39 \\ 
\midrule
LinguGraph-Mistral (7B) & 87.82$\pm$0.88 & 87.47$\pm$0.72 & 93.71$\pm$0.69 & 93.37$\pm$0.52 & LinguGraph-Mistral (7B) & 76.07$\pm$0.53 & 76.02$\pm$0.44 \\
LinguGraph-Llama2 (7B) & 88.19$\pm$0.83 & 88.12$\pm$0.73 & 94.09$\pm$0.78 & 93.55$\pm$0.61 & LinguGraph-Llama2 (7B) & 75.67$\pm$0.52 & 75.60$\pm$0.41 \\
LinguGraph-Llama3 (8B) & \cellcolor[gray]{0.8}91.51$\pm$0.46 & \textbf{91.53$\pm$0.18} & \textbf{95.59$\pm$0.29} & \textbf{95.55$\pm$0.10} & LinguGraph-Llama3 (8B) & \textbf{79.73$\pm$0.18} & \textbf{79.29$\pm$0.56} \\
GCN$_{(Mistral)}$ & 90.04$\pm$0.64 & 89.62$\pm$0.58 & 88.92$\pm$0.71 & 88.47$\pm$0.69 & GCN$_{(Mistral)}$ & 73.55$\pm$0.49 & 73.27$\pm$0.35 \\
GCN$_{(Llama2)}$ & 90.59$\pm$0.71 & 89.62$\pm$0.66 & 88.97$\pm$0.82 & 88.56$\pm$0.71 & GCN$_{(Llama2)}$ & 73.87$\pm$0.22 & 73.87$\pm$0.61 \\
GCN$_{(Llama3)}$ & 90.77$\pm$0.28 & 90.35$\pm$0.37 & 89.76$\pm$0.44 & 89.46$\pm$0.37 & GCN$_{(Llama3)}$ & 74.68$\pm$0.45 & 74.29$\pm$0.32 \\
GAT$_{(Mistral)}$ & 89.85$\pm$0.62 & 89.19$\pm$0.52 & 88.08$\pm$0.62 & 87.53$\pm$0.47 & GAT$_{(Mistral)}$ & 74.72$\pm$0.47 & 74.55$\pm$0.42 \\
GAT$_{(Llama2)}$ & 90.33$\pm$0.67 & 89.72$\pm$0.59 & 87.93$\pm$0.28 & 87.42$\pm$0.36 & GAT$_{(Llama2)}$ & 74.92$\pm$0.14 & 74.48$\pm$0.28 \\
GAT$_{(Llama3)}$ & \cellcolor[gray]{0.8}91.51$\pm$0.35 & \cellcolor[gray]{0.8}91.45$\pm$0.58 & 88.31$\pm$0.76 & 87.93$\pm$0.65 & GAT$_{(Llama3)}$ & \cellcolor[gray]{0.8}75.71$\pm$0.41 & \cellcolor[gray]{0.8}75.06$\pm$0.36 \\
GraphSAGE$_{(Mistral)}$ & 90.59$\pm$0.82 & 89.85$\pm$0.75 & 90.11$\pm$0.69 & 89.77$\pm$0.54 & GraphSAGE$_{(Mistral)}$ & 72.85$\pm$0.42 & 72.87$\pm$0.41 \\
GraphSAGE$_{(Llama2)}$ & 90.22$\pm$0.77 & 89.89$\pm$0.19 & 89.96$\pm$0.50 & 89.67$\pm$0.34 & GraphSAGE$_{(Llama2)}$ & 72.53$\pm$0.61 & 72.42$\pm$0.49 \\
GraphSAGE$_{(Llama3)}$ & \textbf{91.70$\pm$0.51} & 91.08$\pm$0.62 & 90.14$\pm$0.56 & \cellcolor[gray]{0.8}89.96$\pm$0.48 & GraphSAGE$_{(Llama3)}$ & 75.38$\pm$0.38 & 75.22$\pm$0.32 \\
GIN$_{(Mistral)}$ & 89.67$\pm$0.71 & 88.64$\pm$0.54 & 87.83$\pm$0.62 & 87.27$\pm$0.56 & GIN$_{(Mistral)}$ & 73.40$\pm$0.42 & 73.63$\pm$0.34 \\ 
GIN$_{(Llama2)}$ & 90.26$\pm$0.67 & 89.20$\pm$0.48 & 87.73$\pm$0.29 & 87.20$\pm$0.30 & GIN$_{(Llama2)}$ & 73.71$\pm$0.25 & 73.42$\pm$0.10 \\
GIN$_{(Llama3)}$ & 91.33$\pm$0.28 & 91.05$\pm$0.53 & 89.22$\pm$0.79 & 88.87$\pm$0.61 & GIN$_{(Llama3)}$ & 75.64$\pm$0.46 & 75.28$\pm$0.39 \\
\textit{Avg. Dist. Gains} & 4.61\% & 5.22\% & 2.79\% & 2.84\% & \textit{Avg. Dist. Gains} & 3.85\% & 4.22\% \\ 
\bottomrule
\end{tabularx}
\label{tab:result}
\end{table*}

\subsection{Experimental Results and Analyses}

\subsubsection{\textbf{Comparison Performance Analyses}}
To validate the effectiveness of our proposed LinguGKD, we report the accuracy and F1 score of node classification over different datasets to evaluate the performance of various graph learning models. We selected a series of representative single-model graph learning approaches as baselines from three corresponding leaderboards\footnote{\url{https://paperswithcode.com/sota/node-classification-on-cora-60-20-20-random}} \footnote{\url{https://paperswithcode.com/sota/node-classification-on-pubmed-60-20-20-random}} \footnote{\url{https://ogb.stanford.edu/docs/leader_nodeprop/}}, ranging from simple to advanced architecture designs. These include message-passing-based GNNs (such as BernNet \cite{he2021bernnet}, RevGAT \cite{li2021training}, etc.) and more complex graph Transformer models (such as Graphormer \cite{ying2021transformers}, E2EG \cite{dinh2023e2eg}, etc.).
Table \ref{tab:result} shows the node classification results. 

\paragraph{\textbf{Effectiveness Analysis of LinguGraph LLMs Compared to Baselines}}
The experimental results demonstrate that, after graph instruction tuning, the LinguGraph LLMs exhibit strong capabilities in semantic and entity relationship understanding and generalizes well across different datasets. Compared to baseline GNNs, the tuned LinguGraph LLMs achieve state-of-the-art results on all datasets (e.g., LinguGraph-Llama3 (8B) achieved 91.51\%, 95.59\%, and 79.73\% accuracy on the Cora, PubMed, and Arxiv datasets, respectively). Additionally, the performance of the LinguGraph LLM improves with the increase in pre-trained LLM corpus and model parameters. For instance, Llama3-8B, with more model parameters and a larger pre-training corpus than Llama2-7B and Mistral-7B, consistently performs well across multiple datasets. These results strongly support the assertion in \cite{ye2023natural} that LLMs have the potential to become the next generation foundational model for future graph learning.

In graph knowledge distillation frameworks, student models often experience performance degradation compared to teacher models \cite{joshi2021gc,chen2022structure2,samy2023graph2feat}, especially without additional training data. Therefore, selecting an exceptional teacher model is crucial for enhancing the performance of student GNNs. The experimental results above motivate our choice of LLM as the teacher model for graph knowledge distillation, ensuring more effective student GNNs.

\paragraph{\textbf{Performance Gains Analysis of Distilled GNNs Over Baselines}}
From the results, we can observe that the proposed LinguGKD framework can be seamlessly applied to different combinations of teacher LLMs and student GNNs, significantly improving the performance of various student GNNs by extracting knowledge from LLMs without requiring additional training data or modifications to the GNN architecture. 
The average distillation gains of various student GNNs range from 2.79\% in PubMed to 4.61\% in Cora. For instance, on the Cora dataset, the accuracy of the GCN model distilled with LinguGraph-Llama3 (8B) increased to 90.77\%, compared to 86.53\% for the vanilla GCN model. 

Moreover, compared to other advanced GNNs and graph Transformer models, the LinguGKD framework enables basic GNN models to achieve competitive performance with these more complex models through knowledge distillation, and even outperform them in certain scenarios. For instance, the GAT model distilled by LinguGraph-Llama3 (8B) on the Cora dataset achieved 91.51\% accuracy, surpassing the performance of more complex models such as RevGAT (89.11\%) and GraphTransformer (86.42\%). This demonstrates that the LinguGKD framework can significantly enhance the performance of basic GNNs without increasing model complexity, making them more efficient and practical for real-world applications.

This notable performance improvement indicates that the LinguGKD framework can effectively transfer the deep semantic knowledge and complex graph structural understanding from the teacher LLM to the student GNN, thus improving its performance in node classification tasks. 
Furthermore, student GNNs distilled from higher-performing teacher LLMs also achieve higher accuracy and F1 scores in their respective tasks, which indicates significant potential for leveraging the rapid advancements in pre-trained LLMs to continuously enhance the performance of student GNNs, thereby improving the efficiency of applications in production environments.

\paragraph{\textbf{Analysis of Knowledge Distillation Effectiveness Across Different Datasets}}
The effectiveness of knowledge distillation varies across different datasets. For instance, on the Cora dataset, the improvement in student GNNs was substantial, with some models even surpassing the performance of the teacher LLM, such as GraphSAGE$_{Llama3}$ achieving an accuracy of 91.70\%, surpassing its teacher LinguGraph-Llama3 (8B) which had 91.51\%. In contrast, the improvement on the PubMed dataset was relatively modest.
The potential reason is that the LLM's predictions for node classification are primarily derived from the node's textual attributes, meaning it performs better in a structural-free context, with graph structural information contributing less significantly. 
The training of student GNNs used pre-extracted node semantic embeddings provided by the datasets, which lost significant information compared to the raw text inputs used by the LLM, resulting in GNNs relying more on graph structural information for classification. 

Specifically, in the Cora dataset, the overlapping relevance between class labels in the paper titles and abstracts results in nodes that can reasonably belong to multiple categories, making it easier for the teacher LLMs to confuse these categories.
The distilled GNNs, on the other hand, not only retained the advantage of understanding graph structure but also gained the semantic understanding capability of the teacher LLM through the layer-adaptive feature alignment, leading to noticeable performance improvements. 
In contrast, the PubMed dataset has fewer categories, which are well-reflected in the node's textual attributes, allowing the LLM to achieve excellent performance. However, the student GNNs, constrained by the small initial feature dimensions provided by the dataset and significant loss of semantic information, faced limitations in aligning their hierarchical features with those of the LLMs due to extensive cross-category citations, resulting in less pronounced improvements post-knowledge distillation. 

In summary, our proposed LinguGKD framework excels in knowledge distillation, significantly enhancing the performance of student GNNs. The teacher LLM, after graph instruction tuning, demonstrates strong semantic and entity relationship understanding capabilities and generalizes well to different graph datasets, providing robust support for student GNNs. The experimental results also indicate that basic GNN models can achieve performance comparable to, or even surpass, complex graph learning models after distillation through LinguGKD. With the continuous development and performance improvement of pre-trained LLMs, the LinguGKD framework is poised to further enhance the performance of GNNs, thereby advancing the efficiency of applications in production environments.

% ########################################################################
\begin{figure}[t]
    \includegraphics[width=0.41\textwidth]{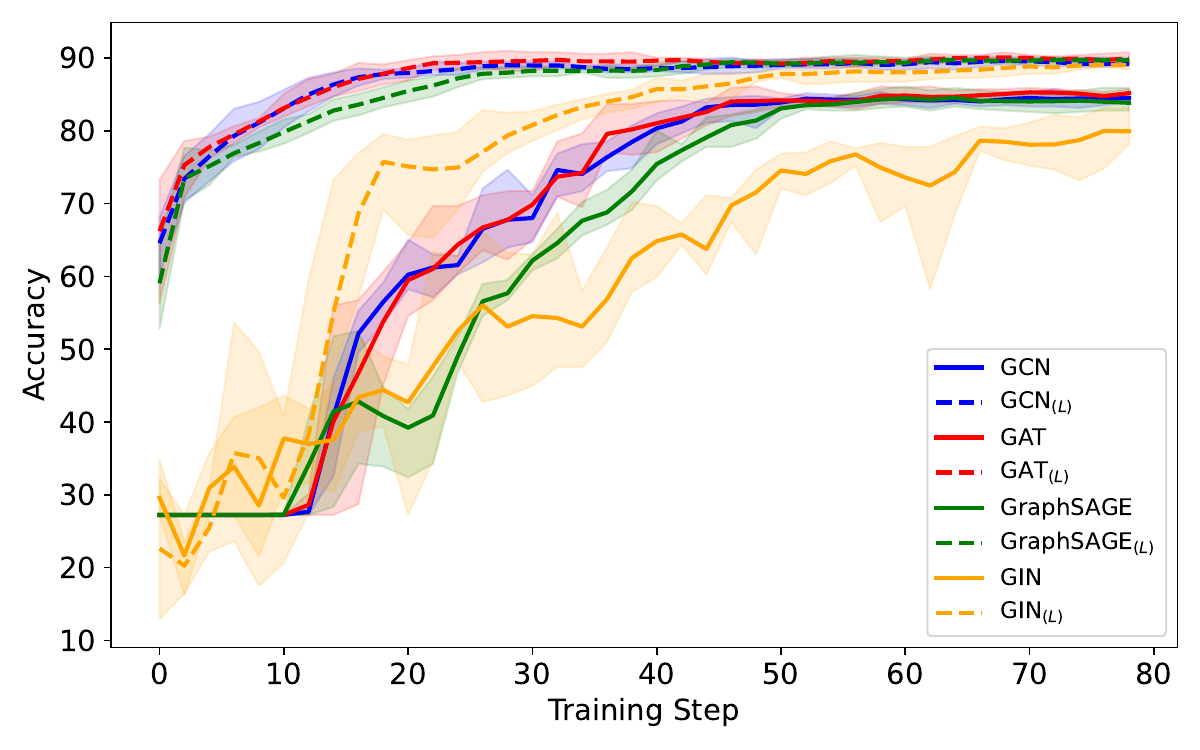}
    \caption{Convergence efficiency of vanilla GNNs and distilled GNNs.}
    \label{fig:convergence}
\end{figure}

\begin{figure}[t]
    \centering
    \includegraphics[width=0.41\textwidth]{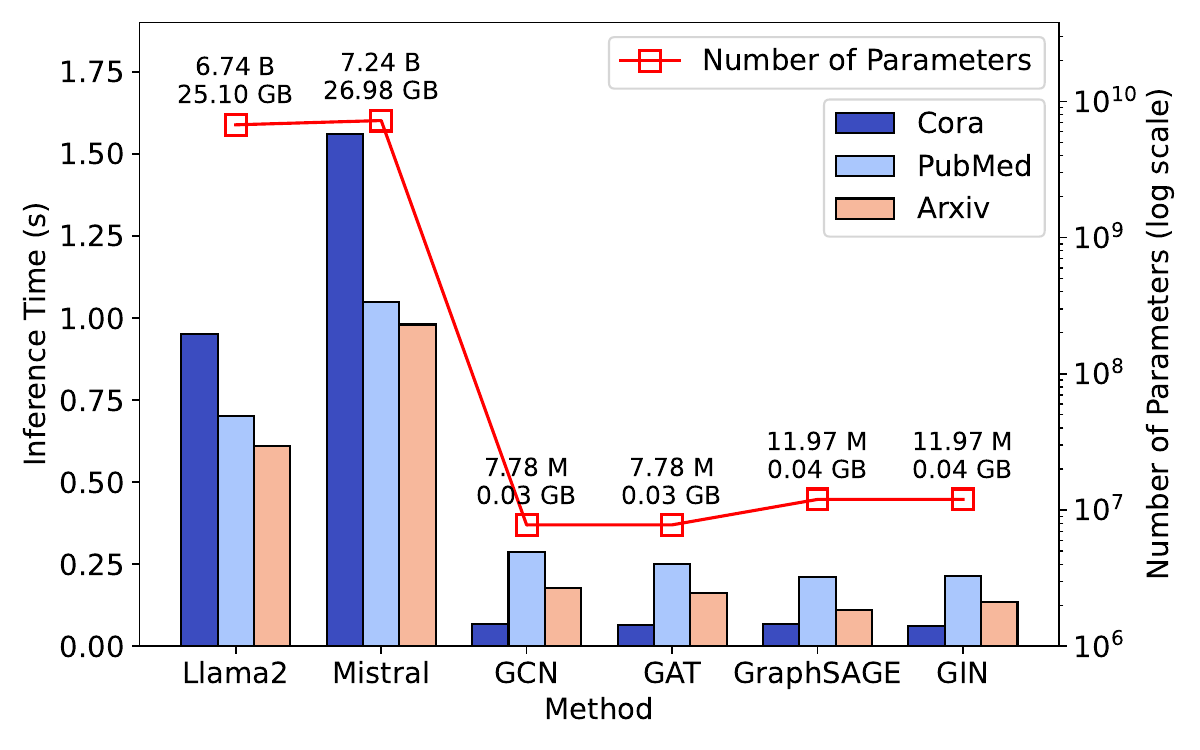}
    \caption{The number of model parameters and inference times.}
    \label{fig:parameters}
\end{figure}
% ########################################################################

\subsubsection{\textbf{Convergence Efficiency of Vanilla GNNs vs. Distilled GNNs}}
Figure \ref{fig:convergence} illustrates that GNNs optimized through knowledge distillation not only achieve higher classification accuracy but also exhibit faster convergence rates. Here, we select LinguGraph-Llama2 as the teacher. The distilled GNNs, distinguished by the subscript $_{(L)}$, quickly reach high accuracy early in the training process, significantly outperforming their undistilled counterparts.

This enhanced convergence is primarily attributed to our joint optimization of knowledge distillation and downstream tasks in the training procedure, in which GNNs are trained to fit the node feature distributions learned by the teacher LLMs, facilitating rapid convergence. Teacher LLMs deliver high-quality node feature representations due to their robust semantic understanding and contextual modeling capabilities. These representations encapsulate complex semantic relationships and extensive contextual information among nodes. When student GNNs learn these feature distributions during the distillation process via the proposed LinguGKD framework, they can significantly reduce the training iterations required to achieve stable high accuracy, resulting in accelerated convergence.

In summary, knowledge distillation not only enhances the classification performance of GNNs but also significantly accelerates model convergence. This advantage renders distilled GNNs more efficient for practical applications, enabling them to complete training in a shorter time while maintaining excellent performance.

\begin{figure*}[t]
    \centering 
    \subfloat[Accuracy on Cora]{
        \includegraphics[width=0.23\textwidth]{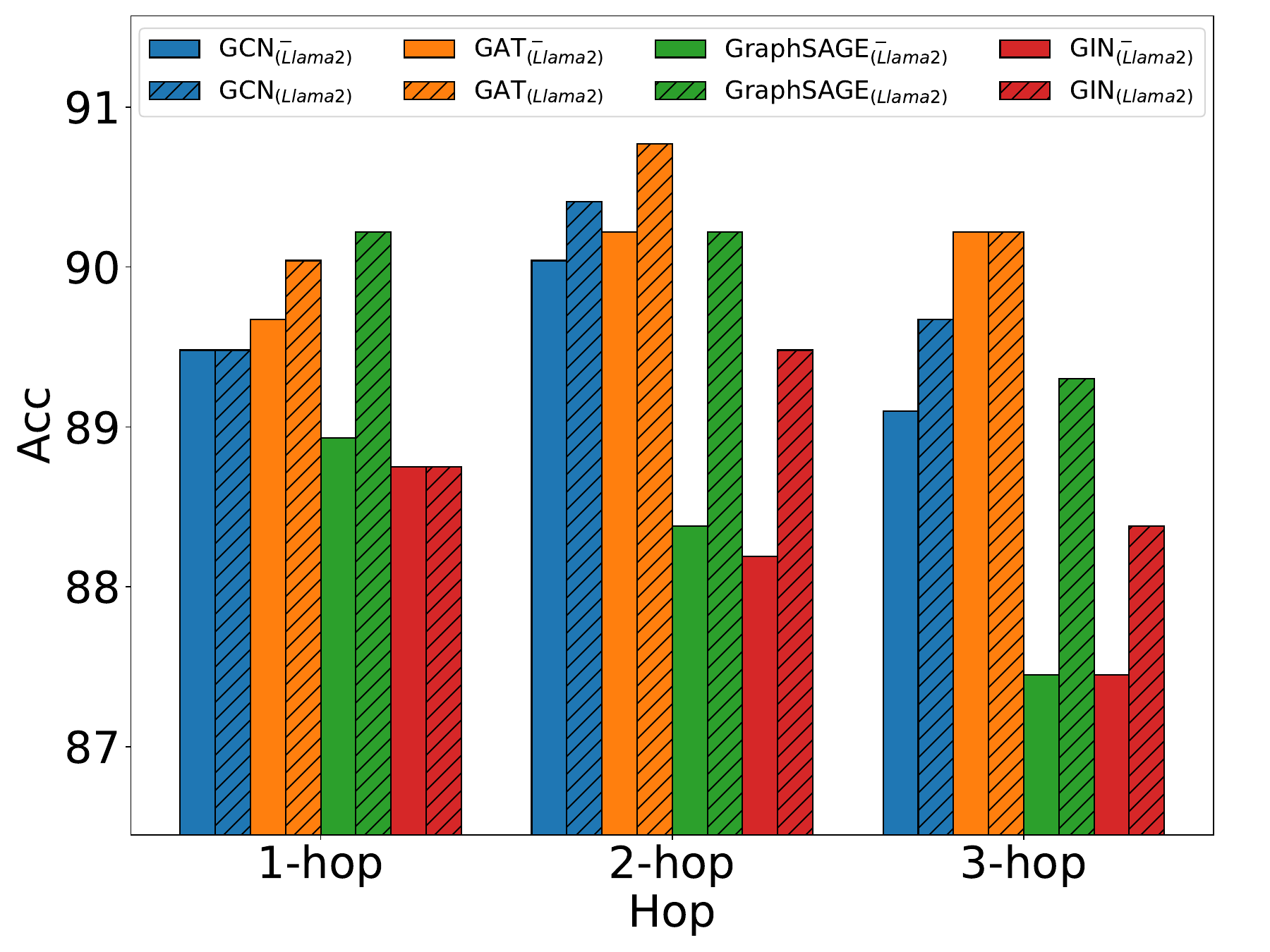}
        \label{fig:ab_cora_acc}
    }
    \hfill
    \subfloat[F1 on Cora]{
        \includegraphics[width=0.23\textwidth]{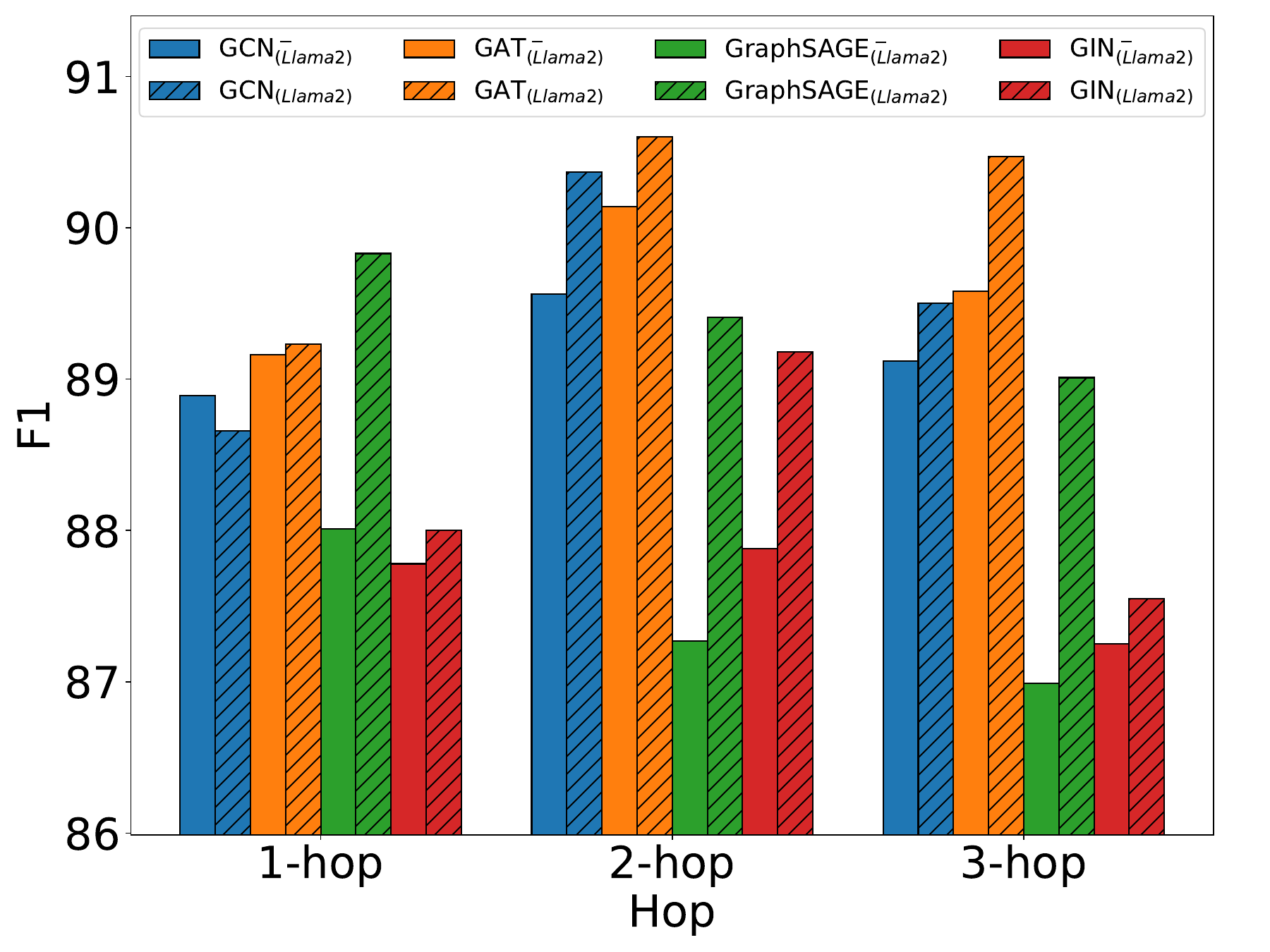}
        \label{fig:ab_cora_f1}
    }
    \hfill
    \subfloat[Accuracy on PubMed]{
        \includegraphics[width=0.23\textwidth]{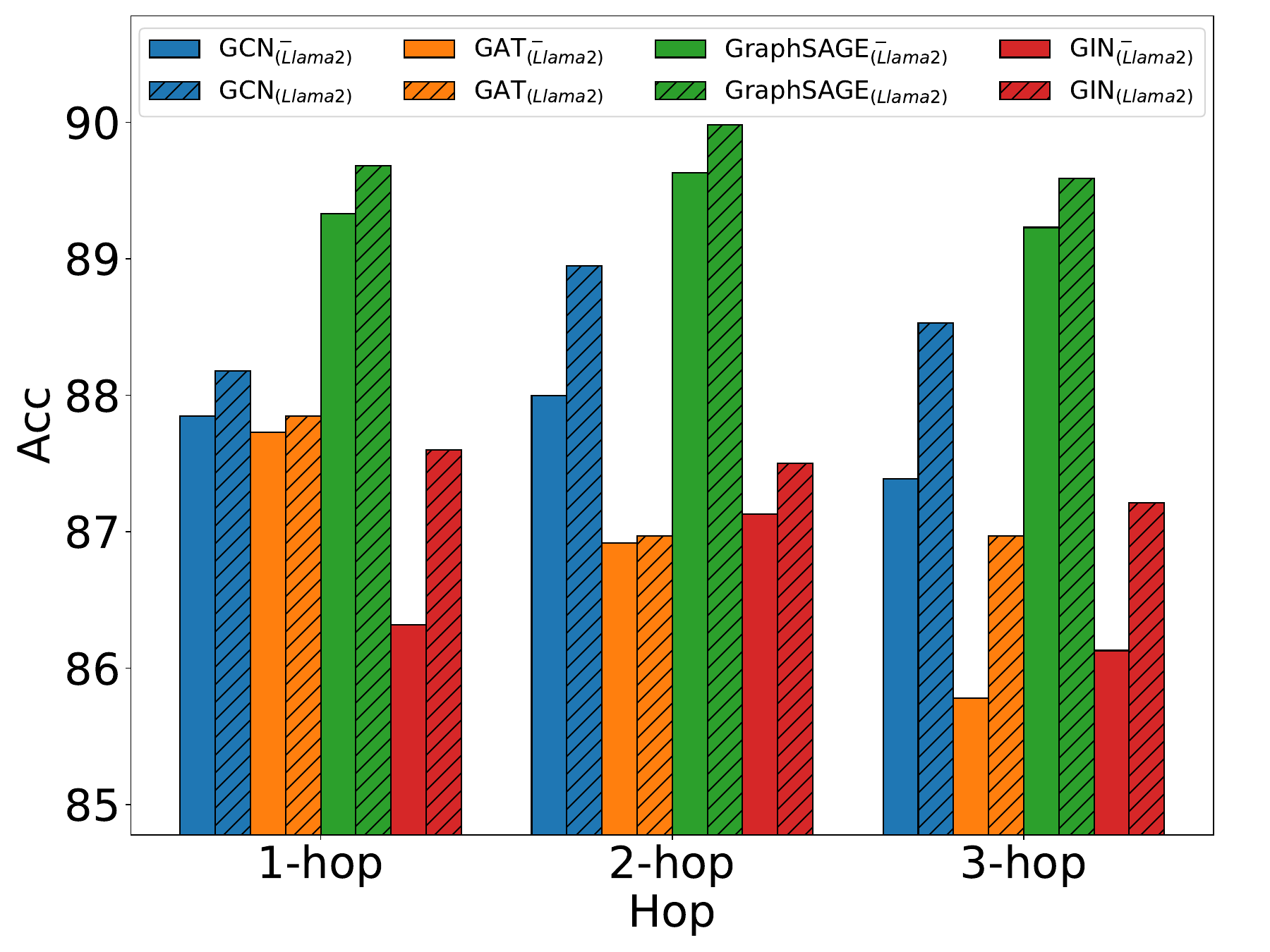}
        \label{fig:ab_pub_acc}
    }
    \hfill
    \subfloat[F1 on PubMed]{
        \includegraphics[width=0.23\textwidth]{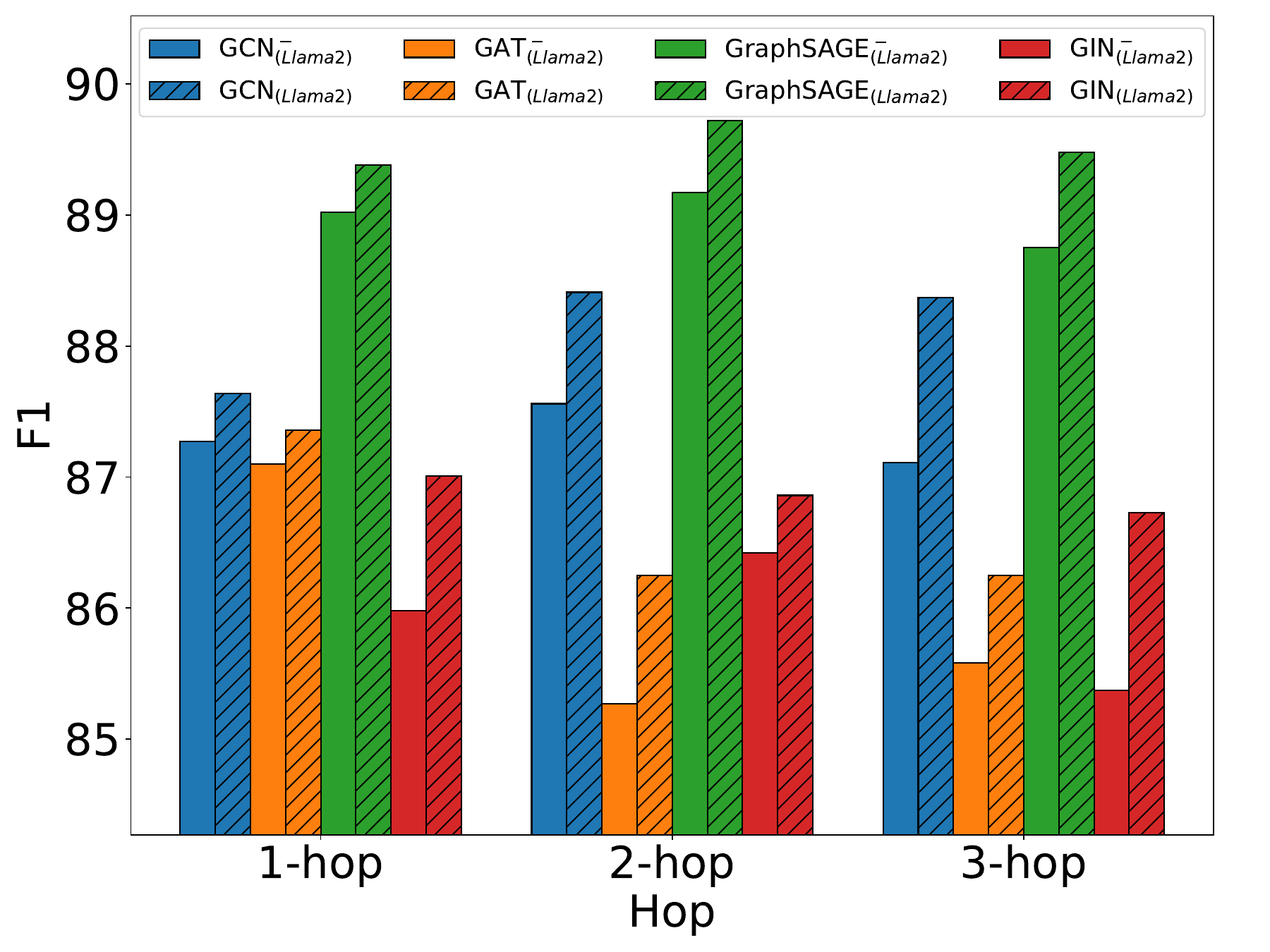}
        \label{fig:ab_pub_f1}
    }
    \caption{Node classification performance comparison of various GNN models distilled with and without layer-adaptive graph knowledge distillation on Cora \& PubMed.}
    \label{fig:ablation}
\end{figure*}

\begin{figure}[t]
    \centering
    \subfloat[Cora]{
        \centering
        \includegraphics[width=0.44\textwidth]{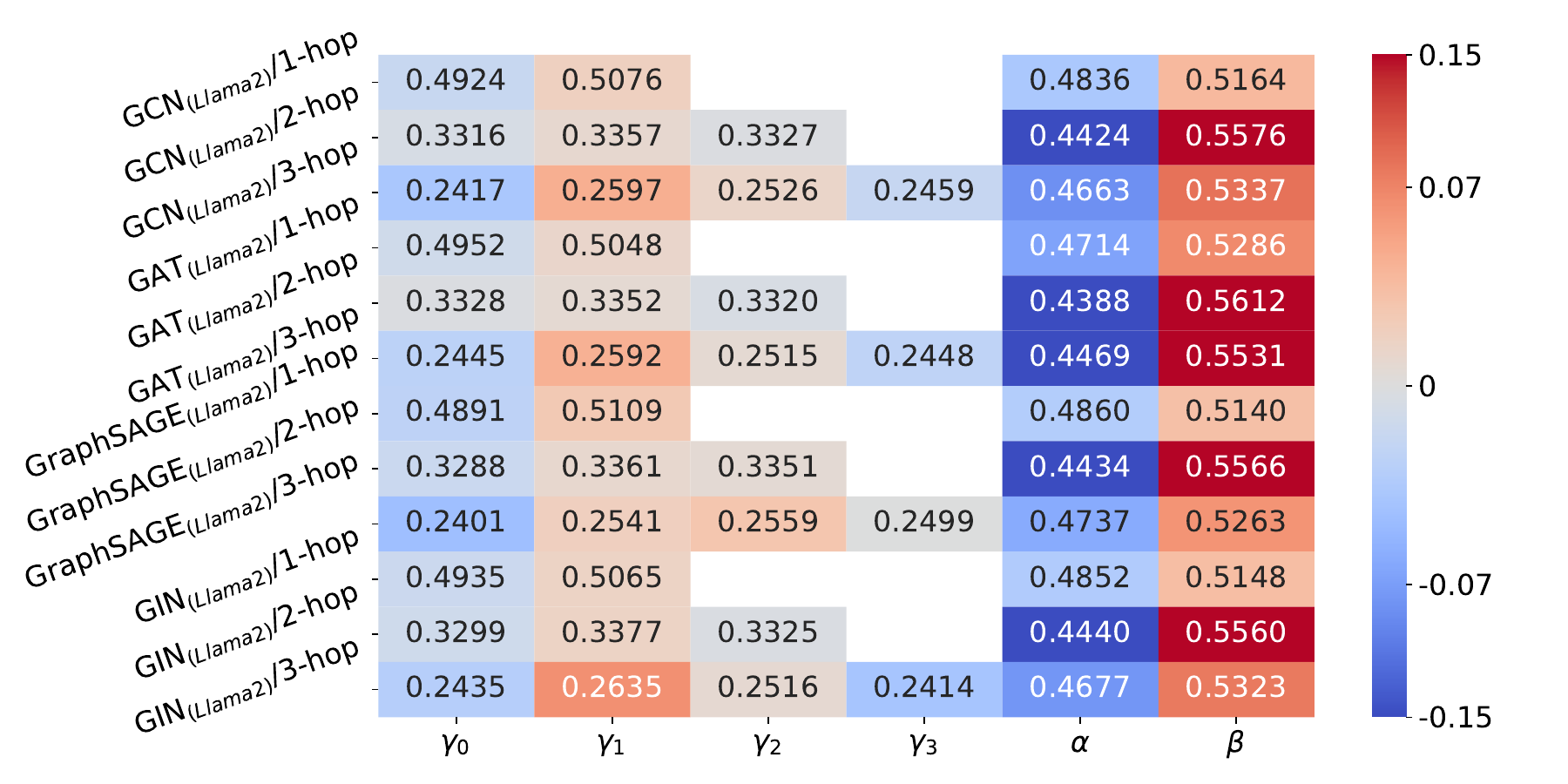}
        \label{fig:weight_cora}
    }
    \hfill
    \subfloat[PubMed]{
        \centering
        \includegraphics[width=0.44\textwidth]{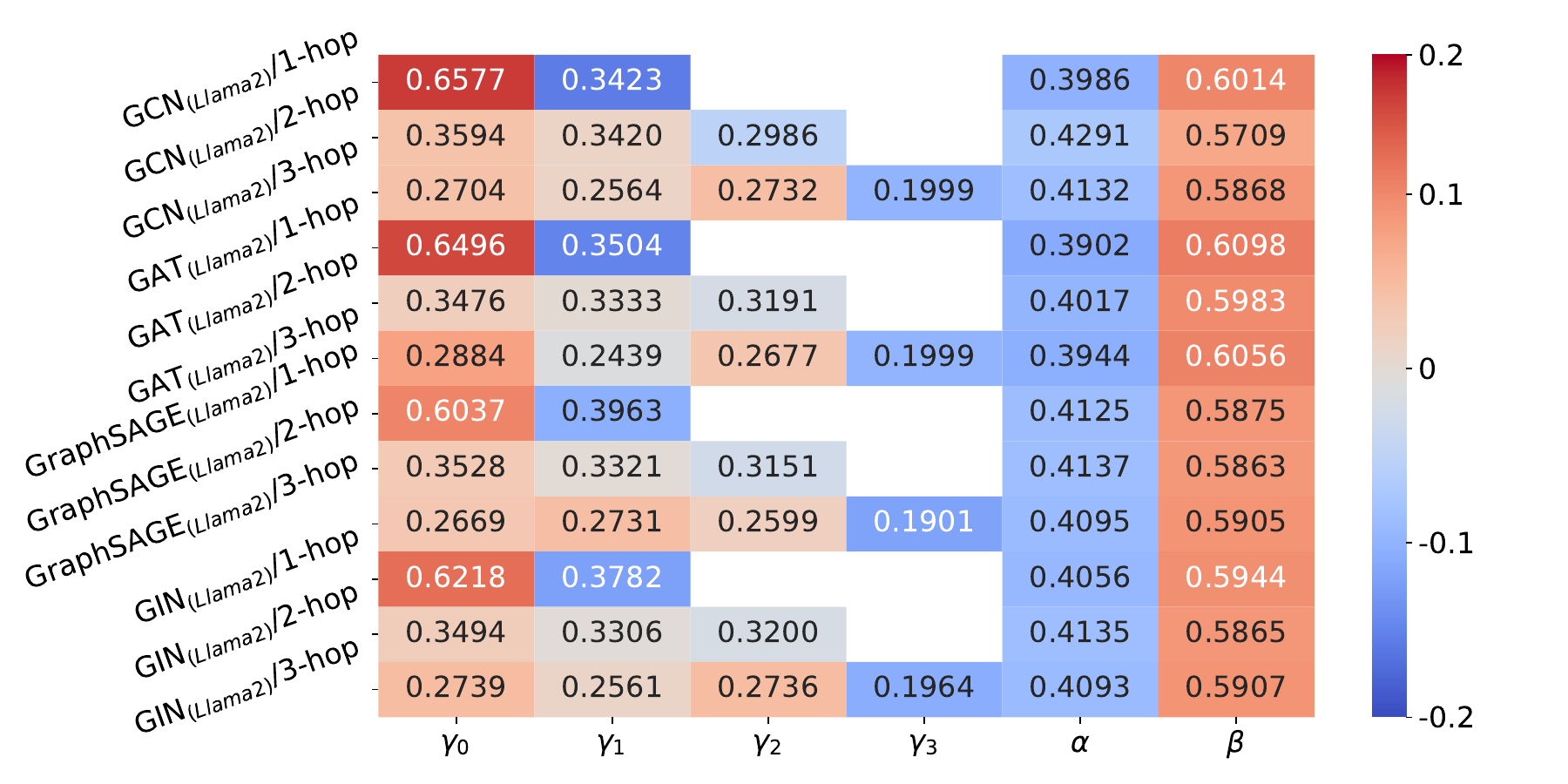}
        \label{fig:weight_pubmed}
    }
    \caption{A heatmap visualization of the layer-adaptive factors $\gamma_l$ and classification-distillation loss weights $\alpha$ and $\beta$ during the training of distilled GNN models on Cora and PubMed datasets. The color depth indicates the differences in various factors and the average value.}
    \label{fig:loss_weights}
\end{figure}

\subsubsection{\textbf{Application Trade-off of LinguGraph LLM vs. Distilled GNN}}
Figure \ref{fig:parameters} shows the differences between teacher LLMs and student GNNs in terms of model parameters, storage requirements, and inference latency. The right y-axis delineates the parameter quantity and storage requirements of various models through a line graph, while the left y-axis showcases the inference latency for these models across different datasets via a bar graph. 

From the figure, we can observe that teacher LLMs have substantially higher parameter counts and storage needs compared to student GNNs. Specifically, Llama2 and Mistral have parameter counts of 6.74B and 7.24B, respectively, while student GNNs have only a few million parameters. In terms of storage, teacher LLMs require over 25GB, whereas student GNNs require just 0.03GB to 0.04GB. Additionally, the inference time for teacher LLMs exceeds 0.5 second on the Cora, PubMed, and Arxiv datasets, whereas the inference time for student GNNs is much smaller.

In practical applications, the choice between using an LLM and a distilled GNN involves several trade-offs. If the application demands the highest possible performance and accuracy, and there are ample computational and storage resources available, deploying a teacher LLM might be preferable. Furthermore, if the application involves tasks that combine natural language processing and graph-based tasks, using a teacher LLM may be essential due to its advanced capabilities in both natural language and graph understanding. However, if the application requires real-time performance or operates under limited resources, a distilled GNN is a more suitable choice. Distilled GNNs not only offer advantages in inference speed and resource consumption but also inherit some capabilities of the teacher LLM through knowledge distillation, striking a good balance between performance and efficiency.

\subsection{Ablation Study}
\begin{table}[t]
\caption{Accuracy of node classification on different datasets for various pre-trained LLMs and their versions after graph instruction tuning.}
\label{tab:tune}
\begin{tabularx}{0.47\textwidth}{@{}lXXX@{}}
\toprule
\textbf{Methods} & \textbf{Cora} & \textbf{PubMed} & \textbf{Arxiv} \\ \midrule
Mistral (7B) & 8.76 & 24.91 & 10.28 \\
LinguGraph-Mistral (7B) & 87.82 & 93.71 & 76.07 \\
Llama2 (7B) & 9.23 & 25.04 & 10.72 \\
LinguGraph-Llama2 (7B) & 88.19 & 94.09 & 75.67 \\
Llama3 (8B) & 14.92 & 30.17 & 15.31 \\
LinguGraph-Llama3 (8B) & 91.51 & 95.59 & 79.73 \\ \bottomrule
\end{tabularx}
\end{table}

\begin{figure*}[t]
    \centering
    \subfloat[Accuracy]{
        \includegraphics[width=0.23\textwidth]{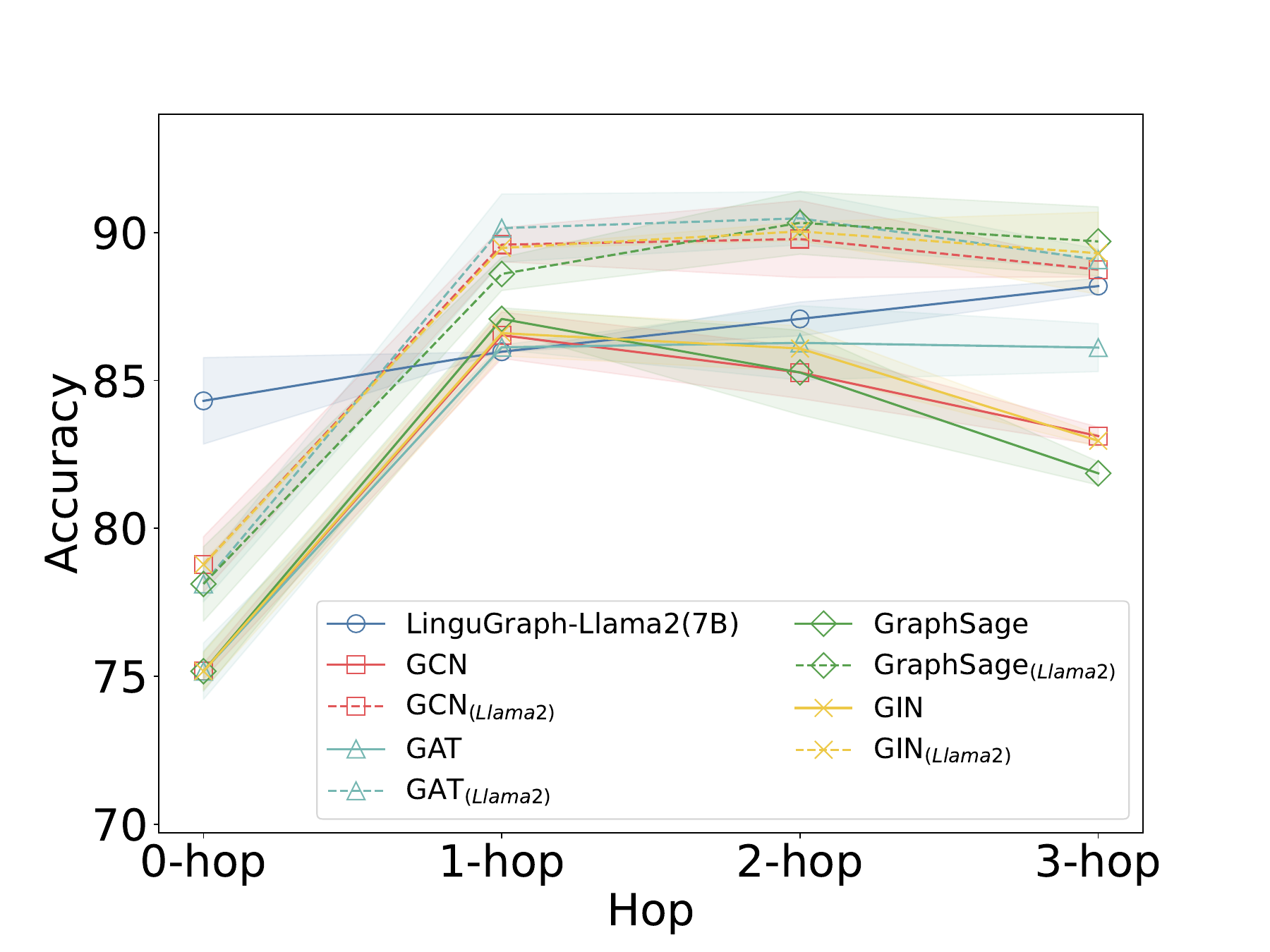}
        \label{fig:accuracy}
    }
    \hfill
    \subfloat[F1]{
        \includegraphics[width=0.23\textwidth]{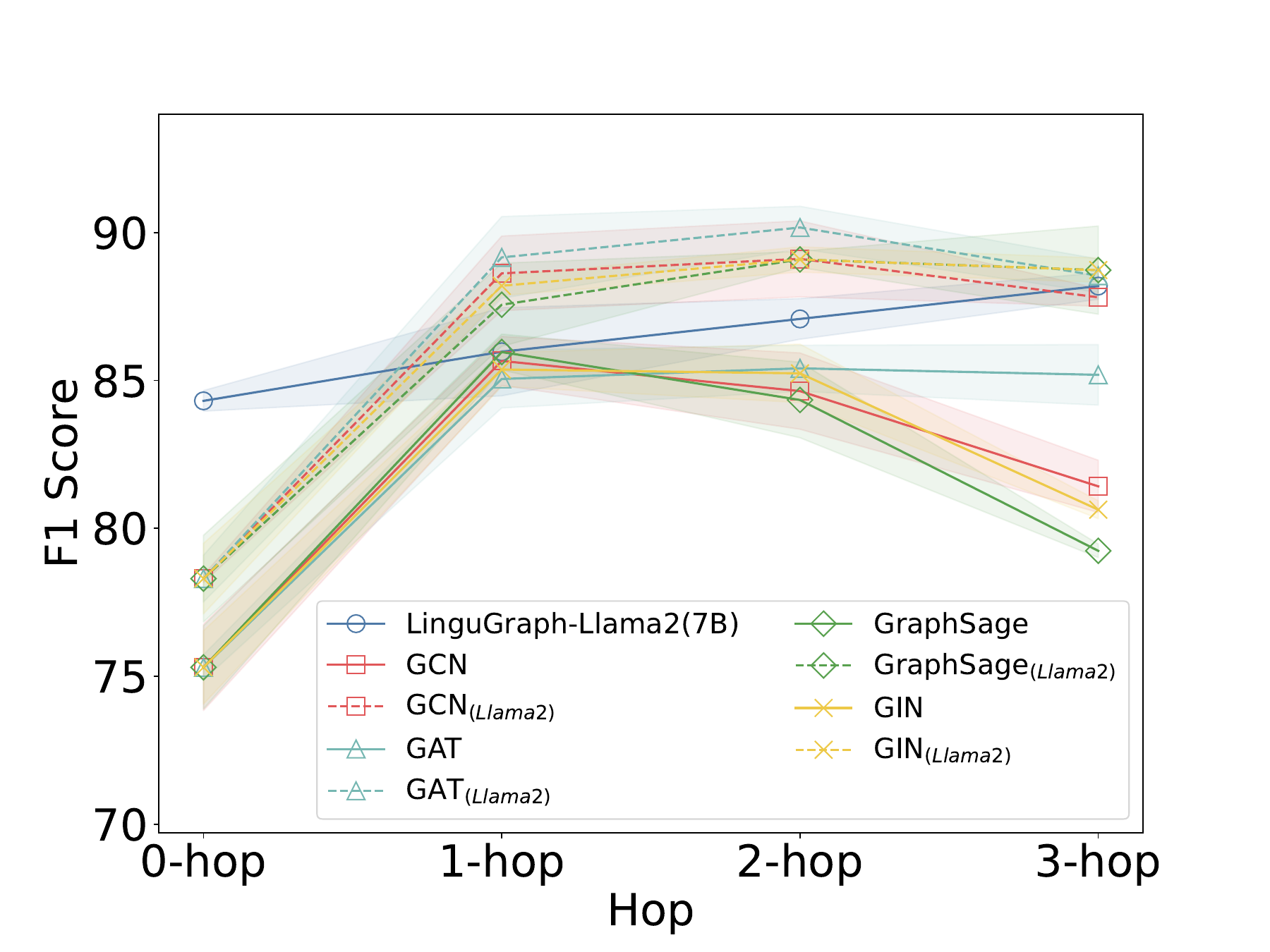}
        \label{fig:f1}
    }
    \subfloat[Accuracy]{
        \includegraphics[width=0.23\textwidth]{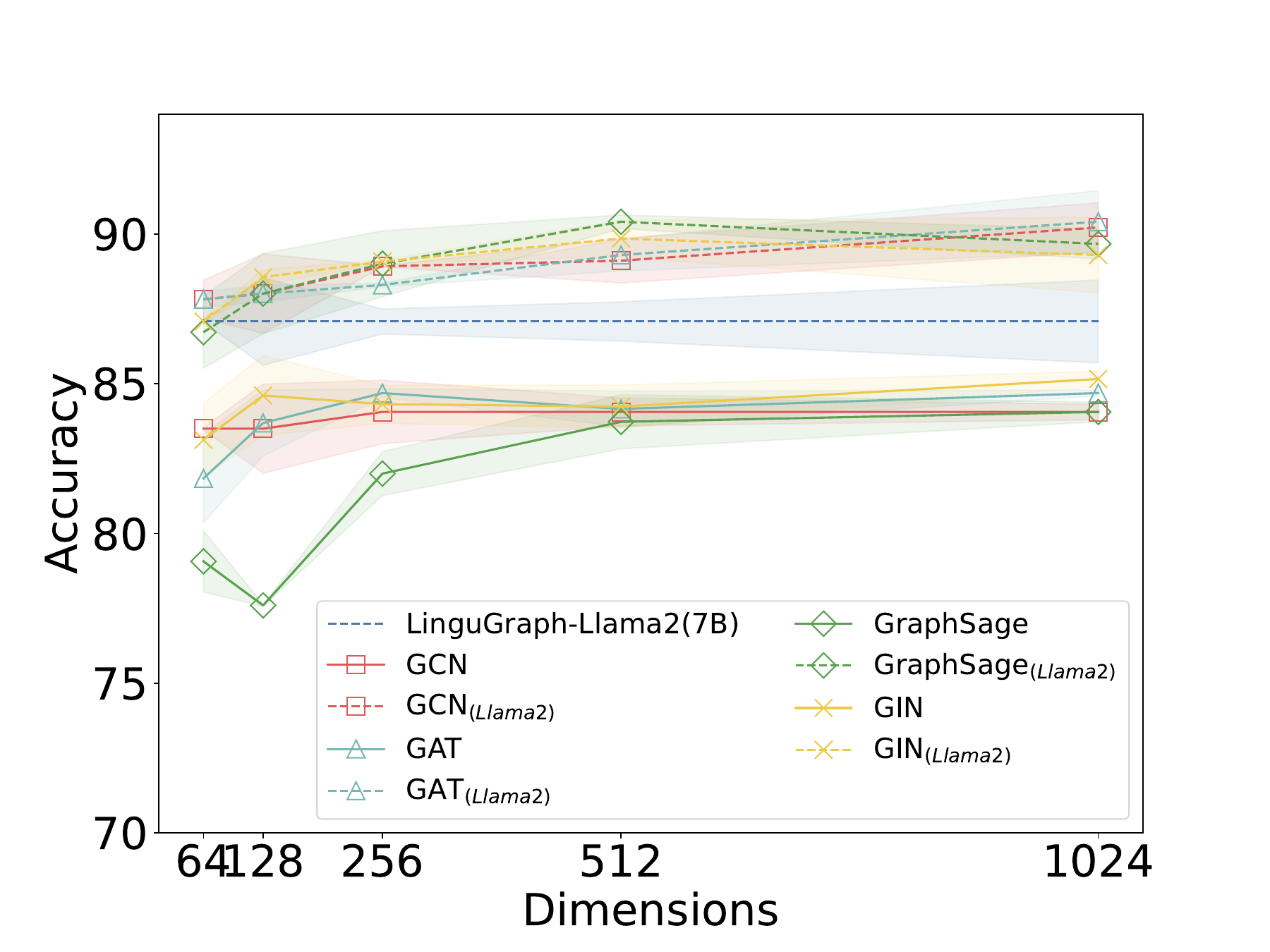}
        \label{fig:accuracy_dim}
    }
    \hfill
    \subfloat[F1]{
        \includegraphics[width=0.23\textwidth]{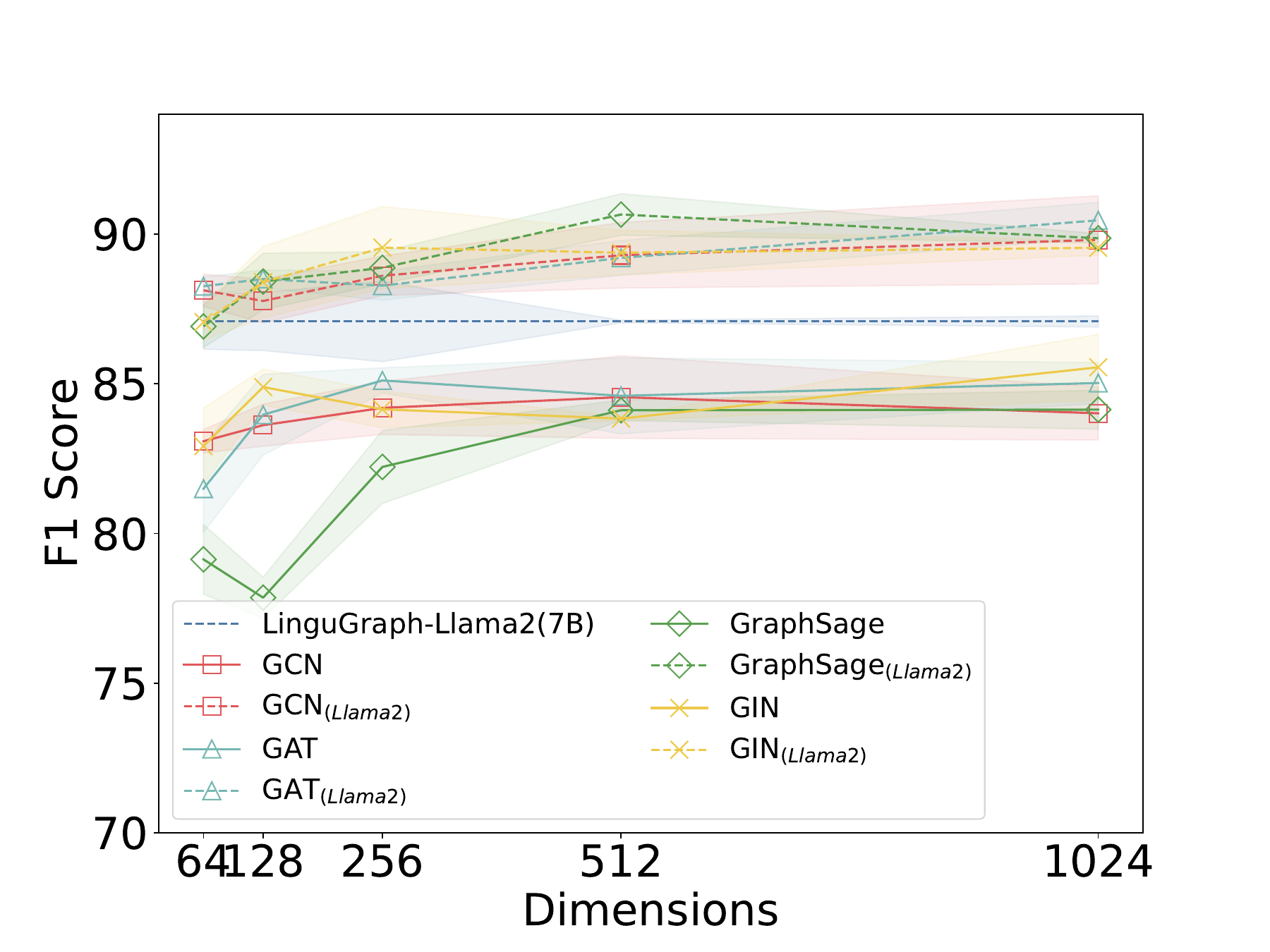}
        \label{fig:f1_dim}
    }
    \caption{The results of performance impacts as the variations of neighbor orders and hidden feature dimensions.}
    \label{fig:dims}
\end{figure*}
\paragraph{\textbf{Necessity Analysis for Graph Instruction Tuning of PLM}}
Table \ref{tab:tune} illustrates the accuracy of node classification on different datasets for various LinguGraph LLMs and their pre-trained versions. The results indicate a substantial improvement in performance across all datasets following the tuning process.

Pre-tuned models exhibited low accuracy scores: around 8-15\% on the Cora dataset, 25-30\% on the PubMed dataset, and 10-15\% on the Arxiv dataset. Since the pre-training of LLMs typically does not include corpora related to graph understanding, they only have a basic understanding of graphs and are prone to issues such as repetition \cite{fu2021theoretical}. However, after graph instruction tuning, their accuracies increased dramatically, with LinguGraph-Mistral (7B) averaging around 86\%, LinguGraph-Llama2 (7B) around 86\%, and LinguGraph-Llama3 (8B) reaching up to 89\% across these datasets.

These results underscore the effectiveness of graph instruction tuning with specifically designed graph instruction prompts in improving the LLMs' ability to understand and classify nodes within graph structures, validating the efficacy of this approach in constructing powerful teacher LLMs for knowledge distillation.

\paragraph{\textbf{Effectiveness Analysis of Layer-Adaptive Contrastive Distillation}}
To validate the effectiveness of the proposed layer-adaptive contrastive distillation, we conducted a set of experiments comparing the impact of using versus not using the layer-adaptive distillation strategy on our proposed LinguGKD framework. We constructed a variant of LinguGKD, named LinguGKD$^-$, which only aligns the last-order node features extracted by the LinguGraph LLM with the features output by the last message-passing layer of the GNN in the distillation space. We used LinguGraph-Llama2 (7B) as the teacher model and conducted comparative experiments on the Cora and PubMed datasets. The node classification performance of GNNs distilled using the LinguGKD framework and the LinguGKD$^-$ variant across different hops are shown in Figure \ref{fig:ablation}, in which GNNs with the $^-$ superscript indicate they were distilled from the LinguGKD$^-$.

From the results, we observe that GNNs distilled with the full LinguGKD framework consistently achieve higher performance on both the Cora and PubMed datasets compared to those distilled with the LinguGKD$^-$ variant. This performance improvement indicates that the layer-adaptive contrastive knowledge distillation strategy effectively enhances the model's ability to leverage multi-hop information and capture nuanced features necessary for accurate node classification. Theoretically, this approach allows the student GNN to progressively align with the hierarchical representations captured by the teacher LLM at different layers, facilitating the transfer of complex, multi-layered knowledge. This alignment helps the student GNN better mimic the teacher LLM’s deep semantic understanding and contextual modeling capabilities, resulting in more robust and accurate node representations. Consequently, student GNNs distilled with the layer-adaptive strategy demonstrate superior performance, validating the robustness and generalizability of our approach.

\paragraph{\textbf{Impact of Layer-Adaptive Distillation and Loss Weights on Model Convergence}}
To further elucidate the contributions of layer-adaptive contrastive knowledge distillation and the interplay between distillation loss and downstream task loss to model convergence during the joint optimization process, we present the layer-adaptive factors $\gamma_l$ and classification-distillation loss weights $\alpha$ and $\beta$ on the Cora and PubMed datasets in Figure \ref{fig:loss_weights}. The heatmaps show that the distillation loss weights ($\beta$) are consistently high across both datasets, indicating the significant role of distillation in accelerating GNN convergence. This trend underscores the importance of contrastive distillation in the training process, as GNNs distilled with higher $\beta$ values converge faster and perform better.

The distribution of layer-adaptive factors $\gamma_l$ varies across different datasets. In Cora, there is a distinct emphasis on first-order neighbor knowledge during distillation, reflected by higher $\gamma_1$ values. This suggests that for Cora, immediate neighborhood information is crucial, likely because the textual attributes and node labels have lower semantic relevance, making structural information vital for effective classification. Conversely, the PubMed dataset emphasizes structure-free features. This suggests that semantic features are more significant in PubMed, where the higher semantic relevance between node textual attributes and labels makes the LLM's semantic understanding crucial. This analysis highlights the critical role of the proposed layer-adaptive contrastive distillation strategy in leveraging the unique structural and semantic characteristics of each dataset.

\subsection{Impact of Hyperparameters on Performance}
Based on the Cora dataset, we investigated how varying hyperparameters, specifically neighbor orders ($k$) and hidden feature dimensions ($d_G$) of GNNs, affects model performance in node classification within TAG. The hidden feature dimensions of the teacher LLMs were kept constant at 4096, as per the original model design.

Figures \ref{fig:accuracy} and \ref{fig:f1} reveal that teacher LLMs consistently outperformed vanilla GNNs, regardless of neighbor hops, highlighting their superior semantic processing ability. Notably, even with no neighbor information (0-hop), LLMs showed a significant edge over GNNs. Under the LinguGKD framework, distilled GNNs surpassed original GNNs in all neighbor hops, demonstrating the effective transfer of multi-hop knowledge from LLMs to GNNs. However, while LLMs benefitted from increasing neighbor orders, GNNs experienced performance declines past the 2-hop mark due to over-smoothing. Distilled GNNs alleviated this issue but faced longer fine-tuning times for teacher LLMs with increased order, leading us to choose a 2-hop setting for a balance of efficiency and effectiveness.

Figures \ref{fig:accuracy_dim} and \ref{fig:f1_dim} provide insights into the impact of hidden feature dimensions on model performance. Vanilla GNNs improved up to a 128-dimension limit, beyond which their performance plateaued. In contrast, distilled GNNs continued to show enhanced performance with higher dimensions, owing to their improved capacity for semantic and structural understanding from LLMs. Consequently, we set the hidden feature dimension at 1024 in our experiments to maximize the benefits from the teacher models.

\section{Related Work}
\label{sec:related_work}
 
\subsection{LLMs based Graph Learning}
Recent advancements in graph learning have been significantly enriched by the integration of LLMs, marking a notable evolution in the field. Research in this domain can be primarily categorized into two distinct approaches: LLM as Enhancer (LaE) \cite{he2023harnessing,chen2024exploring,wei2023llmrec} and LLM as Predictor (LaP) \cite{wang2023can,fatemi2023talk,ye2023natural}, distinguished by their degree of integration with graph-structured data.

The LaE approach enhances node embedding quality in GNNs by leveraging the semantic processing capabilities of LLMs, addressing traditional GNNs' limitations in extracting semantic features from TAGs. For instance, TAPE \cite{he2023harnessing} generates interpretive explanations and pseudo-labels, enriching the graph's textual attributes and fine-tuning a smaller-scale language model to transform textual semantics into robust node embeddings. Similarly, Chen et al. \cite{chen2024exploring} propose the Knowledge Entity Augmentation (KEA) strategy, employing LLMs to generate knowledge entities with textual descriptions, enhancing graph nodes with nuanced semantic information. Other notable methods, such as Qian et al. \cite{qian2023can}, produce semantically-rich interpretations of strings for fine-tuning a compact language model, showing potential in fields like pharmaceutical discovery. Wei et al. \cite{wei2023llmrec} enhance user-item interaction edges in recommendation systems, creating a richer edge dataset and improving system precision.
Several studies have also explored the direct application of LLMs in producing text-based node embeddings for GNNs. The GIANT method \cite{chien2022node} refines language models through a self-supervised learning framework, utilizing XR-Transformers \cite{zhang2021fast} to address multi-label classification challenges in link prediction. Similarly, Duan et al. \cite{duan2023simteg} and Zhu et al. \cite{zhu2023touchup} enhance PLMs using link prediction analogues to improve structural awareness. Huang et al. \cite{huang2023prompt} integrate a graph-tailored adapter at the terminus of PLMs to extract graph-aware node features, generating interpretable node representations. Tan et al. \cite{tan2023walklm} introduce an unsupervised technique for universal graph representation learning, converting random walks on graphs into text sequences for fine-tuning LLMs.

The LaP methodologies utilize LLMs directly for prediction tasks in graph contexts, including classification and inference. 
Recent research leverages LLMs pre-trained on large-scale corpora for encoding graph structures in natural language, enabling direct inference. Studies by Wang et al. \cite{wang2023can} and Fatemi et al. \cite{fatemi2023talk} explore LLMs' ability to process textual descriptions of graphs, highlighting their potential and limitations. Ye et al. \cite{ye2023natural} propose scalable prompting techniques, creating direct relational links between nodes through natural language, outperforming traditional GNNs in node classification tasks across benchmarks.

Collectively, these advancements underscore the potential of LLMs in graph learning, skillfully deciphering both semantic content and complex graph structures, and introducing innovative methodologies to the field.

\subsection{Graph Knowledge Distillation}
In the realm of graph knowledge distillation, the strategic transfer of intricate knowledge from complex teacher models to simpler student models is crucial for enhancing GNNs' effectiveness and efficiency. This technique maintains the student model's lightweight nature while striving to emulate the teacher model's advanced behavior. Predominantly, research in this field is categorized into three areas based on the type of knowledge distilled: output logits \cite{he2022sgkd,ahluwalia2023abkd,wu2020model}, latent features \cite{chen2022structure2,samy2023graph2feat,joshi2021gc}, and graph structure \cite{wu2023pgkd,deng2021graph,yang2023vqgraph}.

Research focusing on output logits in graph knowledge distillation aims at transferring the final output representations from the teacher model to the student model. He et al. \cite{he2022sgkd} proposed the Scalable and Effective Knowledge Distillation Framework (SGKD) for graph representation learning, which includes feature propagation to provide MLPs with graph structure-aware features. Ahluwalia et al. \cite{ahluwalia2023abkd} introduced Attention-Based Knowledge Distillation (ABKD) to compress large GNNs into smaller ones while maintaining accuracy. Wu et al. \cite{wu2020model} developed an approach focused on model extraction attacks on GNNs, demonstrating effective duplication of models with high input-output correlation.

Latent feature distillation involves transferring intermediate representations from the teacher to the student model. Chen et al. \cite{chen2022structure2} proposed a structure-aware MLP student and a structure-mixing distillation strategy to distill knowledge from GNNs into MLPs. Samy et al. \cite{samy2023graph2feat} introduced Graph2Feat that enables inductive link prediction in graph learning through knowledge distillation, showing superior performance in terms of AUC and average precision. Joshi et al. \cite{joshi2021gc} introduced graph contrastive representation distillation (G-CRD), aligning student node embeddings with those of the teacher in a shared representation space to preserve global topology.

Graph structure knowledge distillation focuses on transferring structural information from the teacher to the student model. Wu et al.'s Prototype-Guided Knowledge Distillation (PGKD) \cite{wu2023pgkd} method distills graph structural information from GNNs to MLPs without requiring graph edges. The graph-free knowledge distillation (GFKD) approach by Deng et al. \cite{deng2021graph} models graph topology structures for knowledge transfer without using graph data. Yang et al. \cite{yang2023vqgraph} proposed VQGraph, a framework that transfers structural knowledge from GNNs to MLPs, achieving state-of-the-art performance in GNN-MLP distillation.

All these exceptional works offer valuable references and a robust theoretical foundation for the proposed LinguGKD framework in this paper.

\section{Conclusion}
\label{sec:conclusion}
In this paper, we propose a novel LLM-to-GNN knowledge distillation framework termed LinguGKD, which integrates the semantic understanding capabilities of LLMs with the efficiency and structural insights of GNNs. LinguGKD employs TAG-oriented instruction tuning to train pre-trained LLMs as teacher models and introduces a layer-adaptive contrastive distillation strategy to align and transfer node features between teacher LLMs and student GNNs within a latent space.
Extensive experiments across various LLM and GNN architectures on multiple datasets demonstrates that LinguGKD significantly enhances the predictive accuracy and convergence rate of GNNs without requiring additional training data or model parameters, making them highly practical for deployment in resource-constrained environments. 
Moreover, LinguGKD shows great potential for leveraging advancements in LLM research to continuously augment GNN performance. 

\section*{Acknowledgments}
This work was supported by National Natural Science Foundation of China (No. 62272290, 62172088), and Shanghai Natural Science Foundation (No. 21ZR1400400).

\bibliographystyle{IEEEtran}
\bibliography{ref}

\vspace{-20pt}
\begin{IEEEbiography}[{\includegraphics[width=1in,height=1.25in,clip,keepaspectratio]{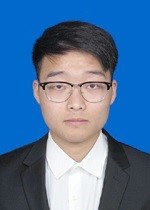}}]{Shengxiang Hu}
is currently a PhD candidate in the School of Computer Engineering and Science at Shanghai University, China. Prior to his ongoing PhD work, he successfully completed his Master's degree in Computer Science and Technology from the same university in 2021. His primary areas of research encompass Quality of Service (QoS) prediction, graph neural networks, and natural language processing. Over the course of his academic career, Hu has contributed significantly to the field through his authorship and co-authorship of 15 scholarly papers. These papers have been published in esteemed international journals and presented at prestigious conferences, such as Knowledge-based Systems, IEEE Transactions on Service Computing, ICSOC, PPSN, etc.
\end{IEEEbiography}

\vspace{-10pt}
\begin{IEEEbiography}[{\includegraphics[width=1in,height=1.3in,clip,keepaspectratio]{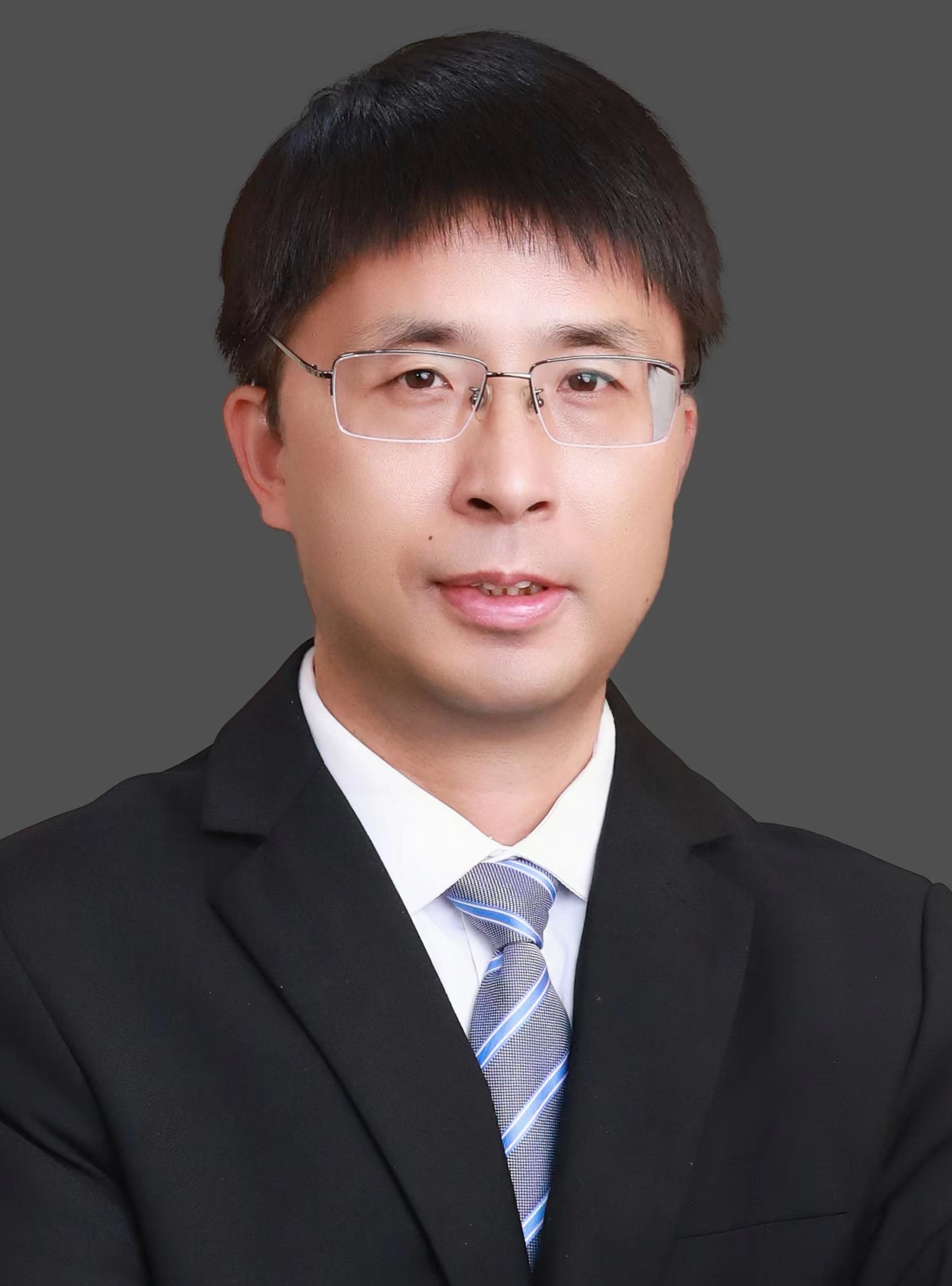}}]{Guobing Zou}
is a full professor and vice dean of the School of Computer Science, Shanghai University, China. He received his PhD degree in Computer Science from Tongji University, Shanghai, China, 2012. He has worked as a visiting scholar in the Department of Computer Science and Engineering at Washington University in St. Louis from 2009 to 2011, USA. His current research interests mainly focus on services computing, edge computing, data mining and intelligent algorithms, recommender systems.  He has published more than 110 papers on premier international journals and conferences, including IEEE Transactions on Services Computing, IEEE Transactions on Network and Service Management, IEEE ICWS, ICSOC, IEEE SCC, AAAI, Information Sciences, Expert Systems with Applications, Knowledge-Based Systems, etc.
% He has published more than 110 papers on premier international journals and conferences, including IEEE Transactions on Services Computing, IEEE Transactions on Network and Service Management, IEEE International Conference on Web Services, International Conference on Service-Oriented Computing, IEEE International Conference on Services Computing, International Journal of Web Services Research, International Journal of Web and Grid Services, AAAI, Information Sciences, Expert Systems with Applications, Knowledge-Based Systems, Applied Intelligence, etc. He served as organization and publicity chair of the International Conference on Service Science, and guest editor of International Journal of Services Technology and Management.
\end{IEEEbiography}
\vspace{-30pt}

\begin{IEEEbiography}[{\includegraphics[width=1in,height=1.25in,clip,keepaspectratio]{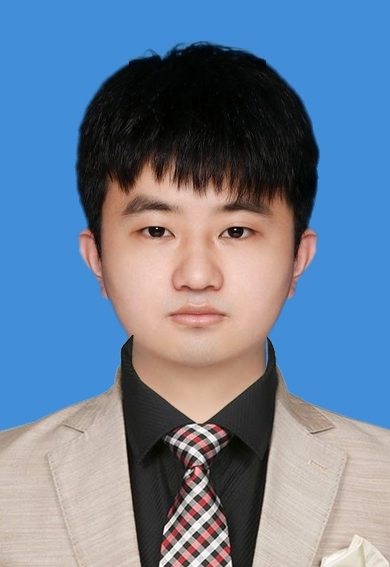}}]{Song Yang}
is currently a PhD candidate in the School of Computer Engineering and Science, Shanghai University, China. He received a Bachelor degree in 2019 and Master degree in 2022 both in Computer Science and Technology at Shanghai University, respectively. His research interests include service computing, edge computing and natural language processing. He has published two papers on IEEE Transactions on Network and Service Management and International Journal of Computational Science and Engineering, respectively. 
% He has led a research and development group to successfully design and implement a service-oriented enterprise application big data platform, which can intelligently classify and recycle, cultivate citizens' habit of throwing recyclables, and produce significant economic and social benefits by providing high QoS.
\end{IEEEbiography}
\vspace{-30pt}

\begin{IEEEbiography}[{\includegraphics[width=1in,height=1.25in,clip,keepaspectratio]{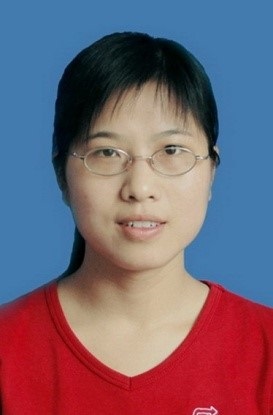}}]{Yanglan Gan}
is a full professor in the School of Computer Science and Technology, Donghua University, Shanghai, China. She received her PhD in Computer Science from Tongji University, Shanghai, China, 2012. Her research interests include bioinformatics, service computing, and data mining. She has published more than 50 papers on premier international journals and conferences, including Bioinformatics, Briefings in Bioinformatics, BMC Bioinformatics, IEEE/ACM Transactions on Computational Biology and Bioinformatics, IEEE Transactions on Services Computing, IEEE Transactions on Network and Service Management, IEEE ICWS, ICSOC, Neurocomputing, and Knowledge-Based Systems.
\end{IEEEbiography}
\vspace{-30pt}

\begin{IEEEbiography}[{\includegraphics[width=1in,height=1.25in,clip,keepaspectratio]{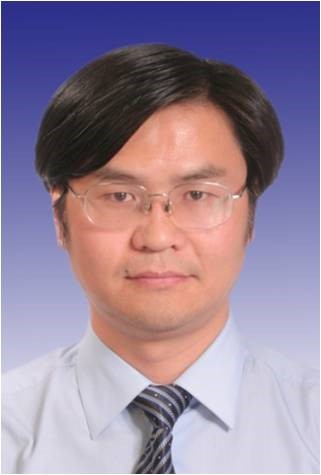}}]{Bofeng Zhang}
is a full professor and dean of the School of Computer and Information Engineering, Shanghai Polytechnic University, Shanghai, China. He received his PhD degree from the Northwestern Polytechnic University (NPU) in 1997, China. He experienced a Postdoctoral Research at Zhejiang University from 1997 to 1999, China. He worked as a visiting professor at the University of Aizu from 2006 to 2007, Japan. He worked as a visiting scholar at Purdue University from 2013 to 2014, US. His research interests include personalized service recommendation, intelligent human-computer interaction, and data mining. He has published more than 200 papers on international journals and conferences. He worked as the program chair for UUMA and ICSS. He also served as a program committee member for multiple international conferences.
\end{IEEEbiography}
\vspace{-30pt}

\begin{IEEEbiography}[{\includegraphics[width=1in,height=1.25in,clip,keepaspectratio]{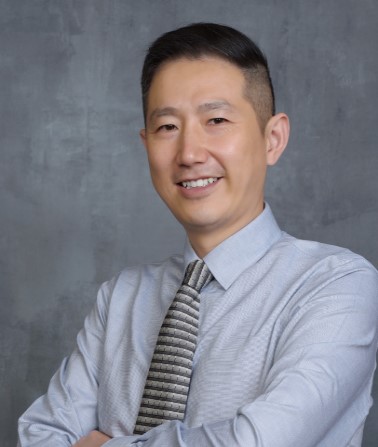}}]{Yixin Chen}
received the PhD degree in computer science from the University of Illinois at Urbana Champaign, in 2005. He is currently a full professor of Computer Science at Washington University in St. Louis, MO, USA. His research interests include artificial intelligence, data mining, deep learning, and big data analytics. He has published more than 100 papers on premier international journals and conferences, including Artificial Intelligence, International Journal of Artificial Intelligence Research, IEEE Transactions on Parallel and Distributed Systems, IEEE Transactions on Knowledge and Data Engineering, IEEE Transactions on Services Computing, IEEE Transactions on Computers, IEEE Transactions on Industrial Informatics, IJCAI, AAAI, ICML, KDD, etc. He won the Best Paper Award at AAAI and a best paper nomination at KDD. He received an Early Career Principal Investigator Award from the US Department of Energy and a Microsoft Research New Faculty Fellowship. He is an Associate Editor for the ACM Transactions on Intelligent Systems and Technology, IEEE Transactions on Knowledge and Data Engineering, and Journal of Artificial Intelligence Research.
\end{IEEEbiography}

\end{document}